# Deep Reinforcement Learning for Dynamic Algorithm Configuration: A Case Study on Optimizing OneMax with the (1+($\lambda$,$\lambda$))-GA

TAI NGUYEN, University of St Andrews, United Kingdom and Sorbonne Université, CNRS, LIP6, France
PHONG LE, University of St Andrews, United Kingdom
ANDRÉ BIEDENKAPP, University of Freiburg, Germany
CAROLA DOERR, Sorbonne Université, CNRS, LIP6, France
NGUYEN DANG, University of St Andrews, United Kingdom

Dynamic Algorithm Configuration (DAC) studies the efficient identification of control policies for parameterized optimization algorithms. Numerous studies leverage Reinforcement Learning (RL) to address DAC challenges; however, applying RL often requires extensive domain expertise. In this work, we conduct a comprehensive study of two deep-RL algorithms–Double Deep Q-Networks (DDQN) and Proximal Policy Optimization (PPO)–for controlling the population size parameter of the $(1 + (\lambda, \lambda))$-GA on OneMax instances. Although OneMax is structurally simple, learning effective parameter control policies for the $(1 + (\lambda, \lambda))$-GA induces a highly challenging DAC landscape, making it a controlled yet demanding benchmark. Our investigation reveals two fundamental challenges limiting DDQN and PPO effectiveness: scalability degradation and learning instability, traced to under-exploration and planning horizon coverage. To address under-exploration, we introduce an adaptive reward shifting mechanism that leverages reward distribution statistics to enhance DDQN exploration. This eliminates instance-specific hyperparameter tuning and ensures consistent effectiveness across problem scales. To resolve planning horizon coverage, we demonstrate that undiscounted learning succeeds in DDQN, while PPO faces fundamental variance issues necessitating alternative designs. We further show that while hyperparameter optimization enhances PPO's learning stability, it consistently fails to identify effective policies across configurations. Finally, DDQN equipped with adaptive reward shifting achieves performance comparable to theoretically derived policies with vastly improved sample efficiency, outperforming prior DAC approaches by orders of magnitude. Taken together, our findings provide insights into the fundamental obstacles faced by standard deep-RL approaches in this challenging DAC setting and highlight the key methodological ingredients required for effective learning.

## 1 Introduction

Evolutionary algorithms (EAs) are well-established optimization approaches, used to solve broad range of problems in various application domains every day. A key factor of EAs' wide adoption in practice is the possibility to adjust their search behavior to very different problem characteristics. To benefit from this versatility, the exposed parameters of an EA need to be suitably configured. Since this tuning task requires substantial problem expertise if done by hand, researchers have developed automated algorithm configuration (AC) tools to support the user by automating this process [Hutter et al. 2009]. AC tools such as IRACE [López-Ibáñez et al. 2016] and SMAC [Lindauer et al. 2022a] are today quite well established and broadly used, especially in academic contexts, with undeniable success in various application domains [Schede et al. 2022].

*Dynamic algorithm configuration (DAC)* extends automated algorithm configuration by learning policies to adapt parameters during optimization runtime rather than identifying fixed parameter values that are used throughout the entire run of the optimization algorithm. While modern

Authors' Contact Information: Tai Nguyen, dptn1@st-andrews.ac.uk, University of St Andrews, St Andrews, United Kingdom and Sorbonne Université, CNRS, LIP6, Paris, France; Phong Le, University of St Andrews, St Andrews, United Kingdom; André Biedenkapp, University of Freiburg, Freiburg, Germany; Carola Doerr, Sorbonne Université, CNRS, LIP6, Paris, France; Nguyen Dang, University of St Andrews, St Andrews, United Kingdom.

evolution strategies like CMA-ES [Hansen 2006] utilize control mechanisms based on current-run data, DAC aims to learn optimal parameter settings across multiple problem instances through transfer learning. Initially explored in [Lagoudakis et al. 2000] and then in [Aine et al. 2008; Andersson et al. 2016; Burke et al. 2013; Karafotias et al. 2012; Kee et al. 2001; Pettinger and Everson 2002; Sakurai et al. 2010; Sharma et al. 2019; Vermetten et al. 2019], the problem of learning control policies through a dedicated training process was formally introduced as DAC in [Adriaensen et al. 2022; Biedenkapp et al. 2020].

Given the large success of reinforcement learning (RL) [Sutton and Barto 1998] in similar settings where one wishes to control state-specific actions, such as in games [Mnih et al. 2013; Silver et al. 2017; Wurman et al. 2022], robotics [Bellemare et al. 2020; Haarnoja et al. 2018b; Lillicrap 2015], or otherwise complex physical systems [Degrave et al. 2022; Kaufmann et al. 2023], it seems natural to address the DAC problem with RL approaches. In fact, the study [Sharma et al. 2019] previously employed a Double Deep-$Q$ Network (DDQN) [Van Hasselt et al. 2016a] method to determine the selection of mutation strategies in differential evolution. Recently, a multi-agent RL approach was used to control multiple parameters of a multi-objective evolutionary algorithm [Xue et al. 2022].

Despite all successes, a number of recent studies also highlight the difficulty of solving DAC problems. Using theory-inspired benchmarks with a known ground truth, the analysis in [Biedenkapp et al. 2022] revealed that a naïve application of RL to DAC settings can be fairly limited, with unfavorable performance in settings of merely moderate complexity. An alternative approach to address DAC problems via sequential algorithm configuration using irace [López-Ibáñez et al. 2016] was suggested in [Chen et al. 2023]. However, this approach requires substantial computational overhead, and its usefulness for more complex settings remains to be demonstrated. The so-called GPS strategy [Levine and Abbeel 2014], used in [Shala et al. 2020] to control the step-size of CMA-ES, requires itself a complex configuration process, causing significant overhead.

In an effort to overcome the challenge of computational demands in Chen et al. [2023] and to leverage RL in DAC, Nguyen et al. [2025a] investigates the learning capabilities of DDQN for dynamically controlling the population size of the (1+($\lambda$,$\lambda$))-GA algorithm optimizing the OneMax instances. This study addresses challenges in the reward function, a critical component in RL training, paving the way to enhance reward design in deep-RL within DAC. Building on the success of DDQN in this context, Nguyen et al. [2025b] extend the (1+($\lambda$,$\lambda$))-GA algorithm solving OneMax with up to four controllable parameters. Their work demonstrates that RL-based DAC significantly outperforms prior methods, including the work by [Chen et al. 2023] using irace—a state-of-the-art static tuning approach. While these studies thoroughly investigate the effectiveness of RL in addressing DAC problems, with a focus on value-based RL, there remains a gap in systematic experimentation with diverse deep-RL approaches and a persistent challenge in applying RL to DAC scenarios.

In this work, we revisit the DAC problem of configuring the $\lambda$ parameter of the (1+($\lambda$,$\lambda$))-GA for optimizing OneMax instances, as initially introduced in Chen et al. [2023] and further enhanced in Nguyen et al. [2025a,b], but contrast two deep-RL solution approaches. There are three major reasons motivating our choice to work with this benchmark. First, it is a well-studied parameter-control benchmark in the theoretical community. In particular, there already exists a policy with the best asymptotic performance for controlling $\lambda$, which we can use to gain valuable insights and evaluate the capability of deep-RL methods in a DAC setting. Second, the benchmark is computationally cheap to run, enabling us to conduct a systematic study of deep-RL algorithms and the impacts of their design choices and hyperparameters. Third, as demonstrated by Chen et al. [2023], the DAC landscape in this benchmark is potentially very challenging. We further confirm this observation via our study with two well-known deep-RL algorithms.

This work is an extension of [Nguyen et al. 2025a], where we explicitly inherit the results of DDQN [Van Hasselt et al. 2016a] in that paper in Sections 3 and 4. The remaining sections present our original contributions, including providing a comprehensive comparison between diverse deep-RL approaches and a more extensive study of methodological improvements for those deep-RL approaches in this DAC setting. We conduct a systematic study of Proximal Policy Optimization (PPO) [Schulman et al. 2017] in the same manner as DDQN in [Nguyen et al. 2025a]. The results, presented in Sections 3 and 4, identify two fundamental issues: scalability degradation as the problem size increases and performance stagnation during training. Our analyses reveal that scalability challenge and learning instability represent more fundamental problems than hyperparameter sensitivity alone, with under-exploration of the policy space emerging as the primary cause.

To tackle the issues of scalability and instability, we adopt *reward shifting* as initially proposed in [Sun et al. 2022] and successfully applied in [Nguyen et al. 2025a,b], which introduces a bias term into the straightforward reward functions, particularly for DDQN. Beyond purely applying reward shifting, we develop a new adaptive bias adjustment mechanism that increases the exploration by leveraging reward value statistics, eliminating the need of instance-specific bias tuning (Section 5). Compared with the adaptive technique introduced in [Nguyen et al. 2025a], our adaptive shifting mechanism effectively captures the exponential increase in runtime as the problem size grows. Comprehensive analyses demonstrate that the newly proposed adaptive shifting offers enhanced learning performances with respect to learning stability and scalability across problem sizes tested.

Section 6 presents the *planning horizon coverage* hypothesis, which states that we empirically validate in our setting, where the analysis reveals that using a conventional discount factor consistently reduces performance when episode lengths differ significantly in the DAC environment. We demonstrate that undiscounted learning (the discount factor $\gamma = 1$) resolves this issue for DDQN, though PPO encounters fundamental variance problems that may necessitate different approaches.

In order to systematically explore alternative solutions for PPO, we conduct an extensive hyperparameter optimization study. This involves identifying key hyperparameters of PPO within the DAC environment and applying the hyperparameter optimization tool SMAC [Hutter et al. 2011] to determine the optimal hyperparameter configuration, as outlined in Section 7. While hyperparameter optimization enhances learning stability, our analysis reveals that PPO inherently struggles to escape local optima. Finally, in Section 8, we show that DDQN policies trained with our adaptive reward shifting can achieve a speed-up of several orders of magnitude to reach the performance of the theory-derived policy compared to the DAC method based on IRACE proposed in [Chen et al. 2023].

**Reproducibility of our results and data availability.** Source code and project data are publicly available at [Anonymous-Github 2025].

## 2 Background

DAC problems are modelled as Markov Decision processes (MDPs) [Bellman 1957]. An MDP $\mathcal{M}$ is a tuple $(\mathcal{S}, \mathcal{A}, \mathcal{T}, \mathcal{R})$ with state space $\mathcal{S}$, action space $\mathcal{A}$, transition function $\mathcal{T}\colon \mathcal{S} \times \mathcal{A} \times \mathcal{S} \to [0, 1]$, and reward function $\mathcal{R}\colon \mathcal{S} \times \mathcal{A} \to \mathbb{R}$. The transition function gives the probability of reaching a successor state $s'$ when playing action $a$ in the current state $s$, thus describing the dynamics of the system. The reward function further indicates if such a transition between states is desirable or if it should be avoided and is crucial for learning processes that aim to learn policies that are able to solve the MDP. In order to describe instance-dependent dynamics and enable learning across multiple instances $i \sim \mathcal{I}$, DAC problems are described as contextual MDPs (cMDPs) [Hallak et al. 2015]. Contextual MDPs extend the MDP formalism through the use of *contextual information* that describes how rewards and transitions differ for different instances while sharing action and state spaces. Consequently, a cMDP $\mathcal{M} = \{\mathcal{M}_i\}_{i \sim \mathcal{I}}$ is a collection of MDPs with shared state and action spaces, but with individual transition and reward functions.

In DAC, a state space represents the algorithm's behavior through its internal statistics during execution, providing necessary context. The action space encompasses all potential parameter configurations. While transition and reward functions are typically unknown and complex to approximate, RL has proven effective for DAC [Adriaensen et al. 2022]. During an offline learning phase, an RL agent interacts with the algorithm being tuned across multiple episodes, each terminating at a goal state or step limit. The agent observes the current state $s_t$, selects action $a_t$, transitions to state $s_{t+1}$, and receives reward $r_{t+1}$. These interactions enable the agent to evaluate states and determine an optimal policy $\pi\colon \mathcal{S} \rightarrow \mathcal{A}$ maximizing expected rewards. While some RL approaches explicitly learn transition models for planning or policy optimization (model-based RL), many widely used modern methods are model-free. These typically either estimate state values $\mathcal{V}\colon \mathcal{S} \rightarrow \mathbb{R}$, state-action values $Q\colon \mathcal{S} \times \mathcal{A} \rightarrow \mathbb{R}$ (e.g. deep $Q$-learning) or directly optimize parameterized policies (policy-gradent-based RL) with auxiliary value functions (actor-critic RL).

## 2.1 Double Deep Q-Network

$Q$-learning [Watkins and Dayan 1992], one of the most widely adopted approaches, aims to learn a $Q$-function that associates each state-action pair with its expected cumulative future reward when taking action $a$ in state $s$. This function is learned through error correction principles. For a given state $s_t$ and action $a_t$, the corresponding $Q$-values $Q(s_t, a_t)$ is updated using temporal differences (TD). A temporal difference describes the prediction error of a $Q$-function with respect to an observed true reward $r_t$ as $TD(s_t, a_t) = r_t + \gamma \max Q(s_{t+1}, \cdot) - Q(s_t, a_t)$, where $\gamma$ is the *discounting factor* that determines how strongly to weigh future rewards in the prediction. For example, with $\gamma = 0$, the temporal difference would describe the error in predicting immediate rewards without the influence of potential future rewards. The estimate of the $Q$-value can then simply be updated using temporal differences as $Q(s_t, a_t) \leftarrow Q(s_t, a_t) + \alpha TD(s_t, a_t)$ with $\alpha$ giving the *learning rate*. A policy can then be defined by only using the learned $Q$-function as $\pi(s) = \arg\max_{a \in \mathcal{A}} Q(s, a)$. To ensure that the state space is sufficiently explored during learning, it is common to employ $\epsilon$-greedy exploration, where with probability $\epsilon$ an action $a_t$ is replaced with a random choice.

Mnih et al. [2013] introduced deep $Q$-networks (DQN) which models the $Q$-function using a neural network and demonstrated its effectiveness in learning $Q$-functions for complex, high-dimensional state spaces, such as video game frames. However, Van Hasselt et al. [2016a] identified that using a single network for action selection and value prediction for computing TDs often creates training instabilities due to overestimation. They proposed to address this issue by using two copies of the network weights: one for selecting the maximizing action and another for value prediction. The second set of weights remains static for brief periods before being updated with the values of the first set. This *double deep* $Q$-network (DDQN) typically reduces overestimation bias and thereby stabilizes learning. This advantage has also helped establish DDQN as one of the most widely used solution approaches in DAC [Li et al. 2023; Ma et al. 2025; Sharma et al. 2019].

## 2.2 Proximal Policy Optimization

Policy-based reinforcement learning directly learns a policy $\pi(a|s;\theta)$ that parameterizes the probability of taking action $a$ in state $s$. A foundational algorithm in this category is REINFORCE [Williams 1992], which uses the policy gradient theorem to update the policy parameters. However, estimating the gradient can be highly uncertain because it relies directly on the returns along the trajectory. To reduce the high variance in REINFORCE, a baseline function is often used, leading to the actor-critic framework [Sutton and Barto 1998], where an *advantage* function [Baird 1993] helps stabilize learning: $A(a_t, s_t) = Q(a_t, s_t) - V(s_t)$. The advantage function estimates how much better (or worse) taking action $a_t$ in state $s_t$ is compared to the average action the policy would take in that state. This function is particularly helpful in making the reinforcement learning agent focus its updates on meaningful action choices rather than the total rewards. Building upon this, Proximal Policy Optimization (PPO) [Schulman et al. 2017] has become a state-of-the-art algorithm for its stability and sample efficiency. PPO constrains the policy update within a small region around the old policy using a clipped surrogate objective:

$$L(\theta) = \mathbb{E}_t \left[ \min \left( r_t(\theta)\hat{A}_t, \mathrm{clip}(r_t(\theta), 1-\delta, 1+\delta)\hat{A}_t \right) \right], \tag{1}$$

where the ratio $r_t(\theta) = \frac{\pi_\theta(a_t|s_t)}{\pi_{\theta_{\text{old}}}(a_t|s_t)}$ measures how much the new policy $(\pi_\theta(a_t|s_t))$ differs from the old policy $(\pi_{\theta_{\text{old}}}(a_t|s_t))$ for the specific state-action pair $(s_t, a_t)$. When $r_t(\theta) = 1$, the policies are identical; when $r_t(\theta) > 1$, the new policy assigns higher probability to the action; when $r_t(\theta) < 1$, it assigns lower probability. The term $\hat{A}_t$ denotes the approximation of the advantage function. The clipping function $\mathrm{clip}(\cdot)$ ensures that the ratio stays within $[1-\delta, 1+\delta]$. For instance, when the advantage is positive, we want to increase the action probability, but clipping at $1+\delta$ prevents excessive increases. Intuitively, the objective $L(\theta)$ allows beneficial policy changes (improving actions with positive advantages) while constraining harmful changes, maintaining training stability by preventing the policy from changing too rapidly in a single update. A key strength of PPO is its ability to handle both discrete and continuous action spaces seamlessly. In discrete action spaces, the policy $\pi(a|s;\theta)$ outputs a probability distribution over all possible actions, typically using a `softmax` output layer. The agent then samples an action from this categorical distribution. On the contrary, in continuous action spaces, the policy commonly parameterizes a Gaussian distribution $\pi(a|s;\theta) = \mathcal{N}(\mu_\theta(s), \sigma_\theta(s))$, where the neural network outputs both the mean and standard deviation of the action distribution. This flexibility has made PPO widely adopted across diverse domains, from discrete game playing to continuous control tasks (e.g., robotics, autonomous driving), and even for DAC [Guo et al. 2025; Tessari and Iacca 2022; Xu et al. 2024].

### 2.3 Benchmarking Dynamic Algorithm Configuration

DACBench [Eimer et al. 2021] provides a standardized collection of DAC problems, including both artificial benchmarks abstracting algorithm runs and real-world benchmarks from various AI domains. The complexity of DAC problems makes establishing ground truth difficult beyond artificial cases, limiting DACBench's capability to evaluate learned policies on real algorithms. This limitation was highlighted when Benjamins et al. [2024] found that cross-instance policies unexpectedly outperformed instance-specific policies designed as performance upper bounds, potentially due to local optima. This underscores the need for ground-truth benchmarks to better understand DAC solutions. While new artificial benchmarks continue emerging [such as, Bordne et al. 2024], theory-inspired DAC benchmarks [Biedenkapp et al. 2022; Chen et al. 2023; Covini et al. 2025; Nguyen et al. 2025a,b] offer a promising middle ground, using theoretical insights from parameter control to provide optimality ground truth while maintaining real algorithm runs. The LeadingOnes benchmark [Biedenkapp et al. 2022] demonstrated this utility by revealing DDQN-based approaches' effectiveness in learning optimal policies for small action spaces while showing limitations with increased dimensionality.

In [Chen et al. 2023], the OneMax-DAC benchmark was introduced. Here, the goal is to control the parameter $\lambda$ of the (1+($\lambda$,$\lambda$))-GA (Algorithm 1) optimizing instances of the OneMax problem $\{f_z : \{0,1\}^n \to \mathbb{R}, x \mapsto \sum_{i=1}^{n} x_i = z_i\}$, also known as 2-color Mastermind. The parameter $\lambda$ determines the population size of the mutation and crossover phase, the mutation rate $p$, and the crossover bias $c$. Optimally controlling $\lambda$ as a function of the current-best fitness is a well-studied problem in the theory community, for which a policy for configuring $\lambda \in \mathbb{R}$ resulting in asymptotically optimal

**Algorithm 1:** The (1+($\lambda$,$\lambda$))-GA with state space $\mathcal{S}$, discrete portfolio $\mathcal{K} := \{2^i \mid 2^i \leq n \wedge i \in [0..k-1]\}$, and parameter control policy $\pi\colon \mathcal{S} \to \mathcal{K}$, maximizing a function $f\colon \{0,1\}^n \to \mathbb{R}$. $\lfloor \lambda \rceil := \lfloor \lambda \rfloor$ if $\lambda - \lfloor \lambda \rfloor < 0.5$, else $\lceil \lambda \rceil$.

1. $x \leftarrow$ a sample from $\{0,1\}^n$ chosen uniformly at random;
2. **for** $t \in \mathbb{N}$ **do**
3.     $s \leftarrow$ current state of the algorithm;
4.     $\lambda = \pi(s)$;
5.     $p = \lfloor \lambda \rceil / n$; and $c = 1/\lfloor \lambda \rceil$;
6.     **Mutation phase:**
7.     Sample $\ell$ from $\mathrm{Bin}_{>0}(n,p)$;
8.     **for** $i = 1, \ldots, \lfloor \lambda \rceil$ **do** $x^{(i)} \leftarrow \mathrm{flip}_{\ell}(x)$; Evaluate $f(x^{(i)})$;
9.     Choose $x' \in \{x^{(1)}, \ldots, x^{(\lfloor \lambda \rceil)}\}$ with $f(x') = \max\{f(x^{(1)}), \ldots, f(x^{(\lfloor \lambda \rceil)})\}$ u.a.r.;
10.     **Crossover phase:**
11.     **for** $i = 1, \ldots, \lfloor \lambda \rceil$ **do**
12.         $y^{(i)} \leftarrow \mathrm{cross}_c(x, x')$;
13.         **if** $y^{(i)} \notin \{x, x'\}$ **then** evaluate $f(y^{(i)})$;
14.     Choose $y' \in \{y^{(1)}, \ldots, y^{(\lfloor \lambda \rceil)}\}$ with $f(y') = \max\{f(y^{(1)}), \ldots, f(y^{(\lfloor \lambda \rceil)})\}$ u.a.r.;
15.     **Selection and update step:**
16.     **if** $f(y') > f(x')$ **then** $y \leftarrow y'$ **else** $y \leftarrow x'$;
17.     **if** $f(y) \geq f(x)$ **then** $x \leftarrow y$;

linear expected optimization time was derived as $\pi_{\mathrm{cont}}(x) := \sqrt{n/(n-f(x))}$ [Doerr and Doerr 2018; Doerr et al. 2015]. The study of OneMax-DAC in [Chen et al. 2023] highlighted the difficulty of this benchmark. Although a tailor-made approach based on irace was able to find policies that performed on par with theoretical policies, its blackbox and “cascading” nature makes it highly *sample inefficient*. In [Nguyen et al. 2025a], the problem of OneMax-DAC is revisited with the goal of addressing the computational demands of the previous method [Chen et al. 2023], demonstrating RL-based DAC, particularly DDQN as a robust approach for dynamically adjusting the population size parameter. However, they acknowledged that while the RL algorithm naturally fits DAC context, it struggles with the exploration problem, and they proposed adopting reward shifting technique [Sun et al. 2022] as the most effective solution. Reward shifting has further evolved to become one of the three key factors underpinning the success of DDQN in multi-parameter control, as discussed in [Nguyen et al. 2025b], where the objective is to extend OneMax-DAC to adjust up to four controllable parameters. Although these prior works paved the way for research on RL solving theory-derived DAC benchmarks, they largely lack a systematic approach, making it difficult to apply RL-based DAC to other benchmarks.

## 3 Deep-RL for Solving OneMax-DAC

Compared to black-box DAC approaches, such as those based on irace, deep reinforcement learning (deep-RL) algorithms, especially off-policy algorithms like DDQN, are expected to be much more sample-efficient, since the learned policies can be updated *during* every episode (i.e., every algorithm run). This property makes deep-RL and similar approaches appealing for DAC scenarios where each solution evaluation is expensive.

However, deep-RL is commonly known to be difficult to use [see, e.g., Parker-Holder et al. 2022]. This difficulty is particularly pronounced for PPO, which relies on hyperparameters that are highly sensitive to the design choices [Adkins et al. 2024; Andrychowicz et al. 2020; Huang et al. 2022]. In this section, we systematically investigate the learning capabilities of two representative deep-RL algorithms, including DDQN (off-policy) and PPO (on-policy), on the OneMax-DAC benchmark. We demonstrate that a naïve application of these commonly used deep-RL algorithms with straightforward reward function results in limited learning ability. This motivates our subsequent investigation of reward function design and exploration mechanisms.

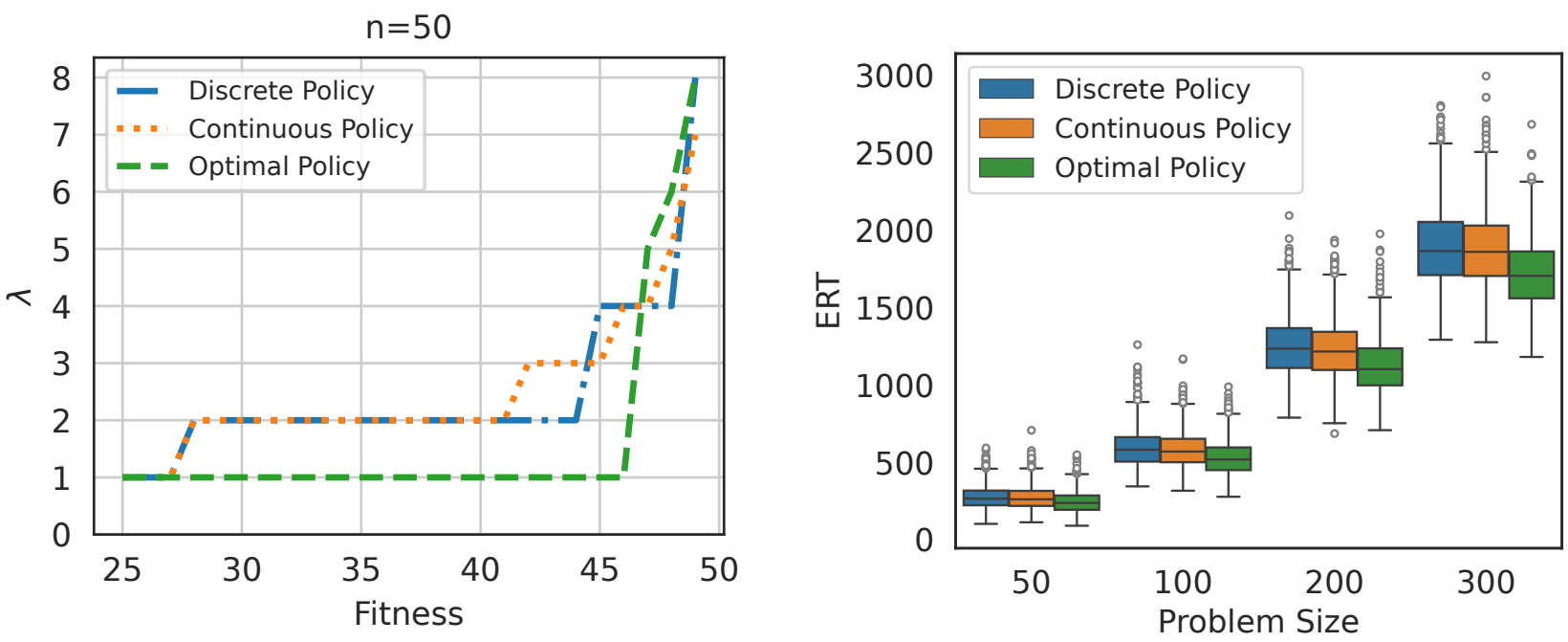

Fig. 1. theory-derived policy, its discretized version, and the optimal policy for $f(x) \geq n/2$ with $n = 50$ (left); their average runtimes across 4 problem sizes over 1,000 runs (right).

**Action Space.** As DDQN is designed to work with a discrete action space, while PPO can handle both continuous and discrete action spaces, we discretize the OneMax-DAC benchmark action space. For a given problem size $n$, we define the set of possible $\lambda$ values that the RL agent can choose from as $\{2^i \mid 2^i \leq n, i \in [0.. \lfloor \log_2 n \rfloor]\}$. With this new action space, we define the following discretized version $\pi_{\texttt{disc}}(x)$ of the theory-derived policy: for a given solution $x$, we choose the $\lambda$ value from the set that is the closest to $\pi_{\texttt{cont}}(x)$.

Figure 1 shows that the difference in performance between $\pi_{\texttt{cont}}$ and $\pi_{\texttt{disc}}$ is marginal (and not statistically significant according to a paired two-sided t-test with a confidence level of 95%). Therefore, an RL agent using this discretized action space should be able to find a policy that is at least competitive with the theory-derived one. Additionally, there is a gap between the optimal policy ($\pi_{\texttt{opt}}$) [Chen et al. 2023] and $\pi_{\texttt{cont}}$ across all problem sizes. Therefore, we aim to propose an RL-based approach to produce a policy closer to $\pi_{\texttt{opt}}$.

**State Space.** Following both theoretical and empirical work on the benchmark [Chen et al. 2023; Doerr and Doerr 2015; Nguyen et al. 2025a,b], we only consider the state space defined by the quality ("fitness") of the current-best solution; in the absence of ground-truth, more complex state spaces are left for future work.

**Reward Function.** The aim of the OneMax-DAC benchmark is to find a policy that minimizes the runtime of the (1+($\lambda$,$\lambda$))-GA algorithm, i.e., the number of solution evaluations until the algorithm reaches the optimal solution. Therefore, an obvious component of the reward function is the number of solution evaluations at each time step (i.e., iteration) of the algorithm.

Additionally, to reduce time collecting samples from bad policies during the learning, following [Biedenkapp et al. 2022], we impose a cutoff time on each run of the (1+($\lambda$,$\lambda$))-GA algorithm, allowing an episode to be terminated even before an optimal OneMax solution is reached. To distinguish the performance between runs terminated due to the cutoff time, denoting by $E_t$ the total number of solution evaluations at time step $t$, and by $\Delta f_t = f(x_t) - f(x_{t-1})$ the fitness improvement between $t$ and $t-1$, we define the reward function as:

$$r_t = \Delta f_t - E_t \tag{2}$$

**Baseline Policies.** We consider three baseline policies in our study, including the theory-derived policy $\pi_{\texttt{cont}}(x)$, its discretized version $\pi_{\texttt{disc}}(x)$, and the (near) optimal policy $\pi_{\texttt{opt}}(x)$ from [Chen et al. 2023].

**Experimental Setup.** We train two deep-RL agents on four OneMax problem sizes, spanning from 50 to 300. For each size, we repeat each RL training 10 times using a budget of 500,000 training steps. The training used a machine equipped with single-socket AMD EPYC 7443 24-Core Processor. We use a single thread for training and parallelize 10 threads for evaluation. A cutoff time of $0.8n^2$ as initially introduced in [Biedenkapp et al. 2022], sufficiently larger than the optimal linear running time, is imposed on each episode during the training.

Following [Biedenkapp et al. 2022; Nguyen et al. 2025a], we utilize a standard hyperparameter configuration of DDQN with $\epsilon$-greedy exploration and $\epsilon = 0.2$. The replay buffer size is set to 1 million transitions. At the beginning of the training process, we sample 10,000 transitions uniformly at random and add them to the replay buffer before learning begins. The Adam optimizer [Kingma and Ba 2015] is used for the training process with a (mini-)batch size of 2,048 and a learning rate of 0.001. To update the $Q$-network, we adopt a discount factor of 0.99 and use the soft target update mechanism with $\tau = 0.01$ to synchronize the online policy and the target policy [Lillicrap et al. 2016]. For the network architecture, we employ a structure comprising two linear layers, each containing 50 hidden nodes, followed by a ReLU activation function [Agarap 2018].

For PPO, we employ the implementation of Stable Baselines [Hill et al. 2018]. We adopt default hyperparameters as follows: a learning rate of 0.0003, 2,048 rollout steps before policy updates, a (mini-)batch size of 64, and 10 training epochs. Additionally, we use a discount factor of $\gamma = 0.99$, a Generalized Advantage Estimation (GAE) [Schulman et al. 2015] coefficient of 0.95 to balance bias and variance in the advantage function, and a clip range of 0.2 for the probability ratio between new and old policies. To enhance learning stability, we always normalize the advantage. For a fair analysis, we also employ an actor and critic networks with the same architecture as DDQN.

**Performance Metrics.** To study the learning performance of each RL training, we record the learned policies at every 2,000 training steps and evaluate each of them with 100 different random seeds. We then measure the performance of each RL training via four complementary metrics: *(1–2) Best policy's performance (ERT and gap):* evaluate top 5 policies across 1000 random seeds, select the best performer, and compute its *expected runtime (ERT)* and *gap relative to baseline policy's ERT*; *(3) Area Under the Curve (AUC):* difference between the learning curve and the baseline performance of $\pi_{\texttt{disc}}(x)$ in Figure 3, where the curve points represent the average runtime of the current policy across 100 seeds; *(4) Hitting rate (HR):* ratio $n_h/n_e$ of policies with runtime within $\mu \pm 0.25\sigma$ of the baseline policy (where $\mu$, $\sigma$ are the ERT and standard deviation) [Biedenkapp et al. 2022; Nguyen et al. 2025a], with hitting points (denoted as $n_h$) determined by counting the number of times the performance of the RL-based policy falls within the shaded area of the discrete policy, as illustrated in Figure 3, and $n_e$ the total number of policies being evaluated. We aim to minimize ERT (gap) and AUC, while maximizing HR.

**Naïve Rewards Fail to Scale.** Figure 2 depicts the gap of the best-learned policies to $\pi_{\texttt{disc}}$ across 10 RL runs using the reward function defined in Equation (2). The performances of DDQN are represented in blue boxes, while those of PPO are depicted in green boxes. The blue boxes highlight the limitation in the DDQN learning scalability: Although the DDQN agent can find policies of reasonable quality in the smallest problem size of $n = 50$ (with an average gap of 3.23%), the gap increases significantly with $n$. From $n = 100$ onward, the agent is no longer able to get close to the baseline during the whole learning process. The same challenge is evident in the green boxes of PPO, but it is more severe. The learning process of PPO cannot even handle the smallest problem size of 50.

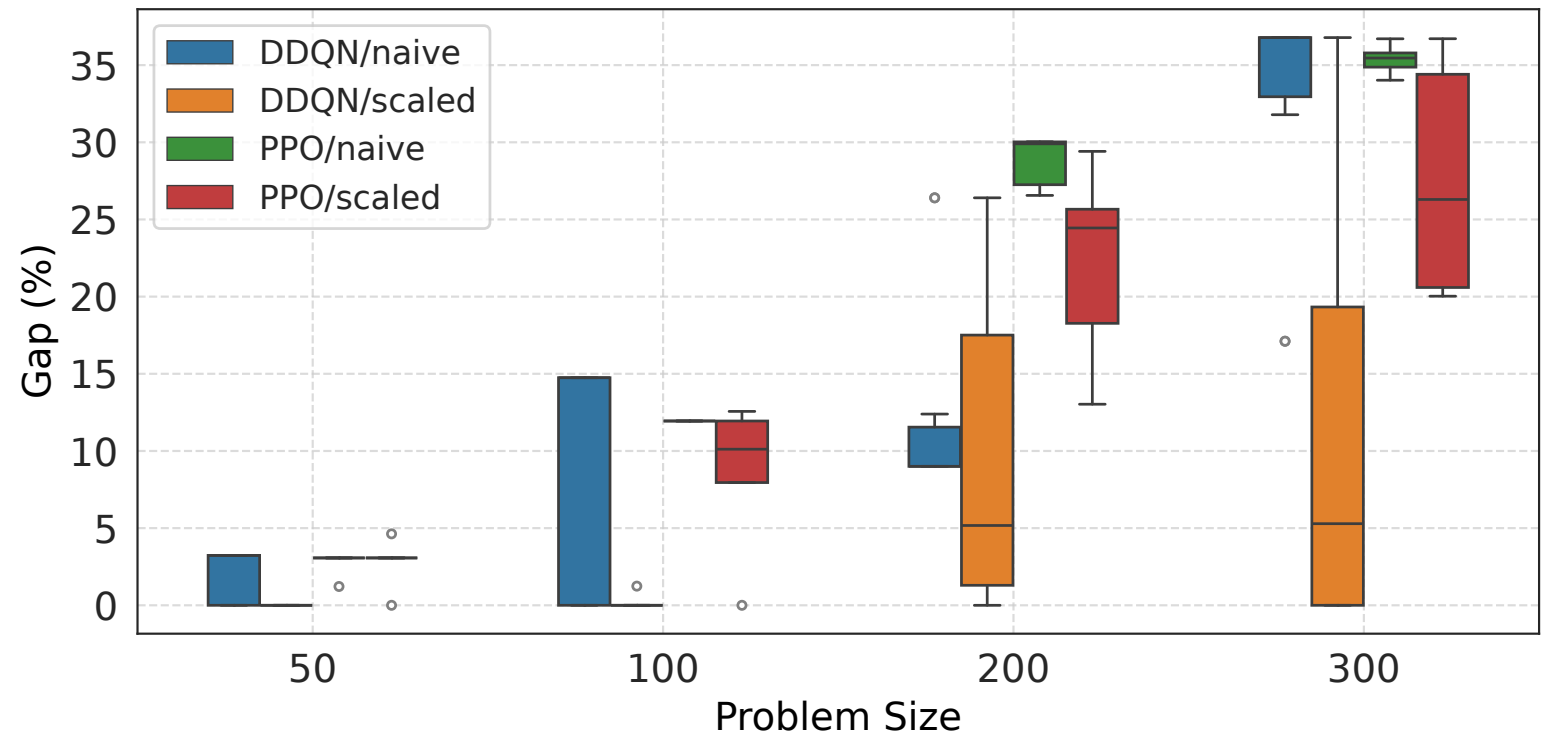

Fig. 2. Performances of DDQN and PPO (as gap to $\pi_{\mathtt{disc}}$ ) using naïve and scaled reward functions.

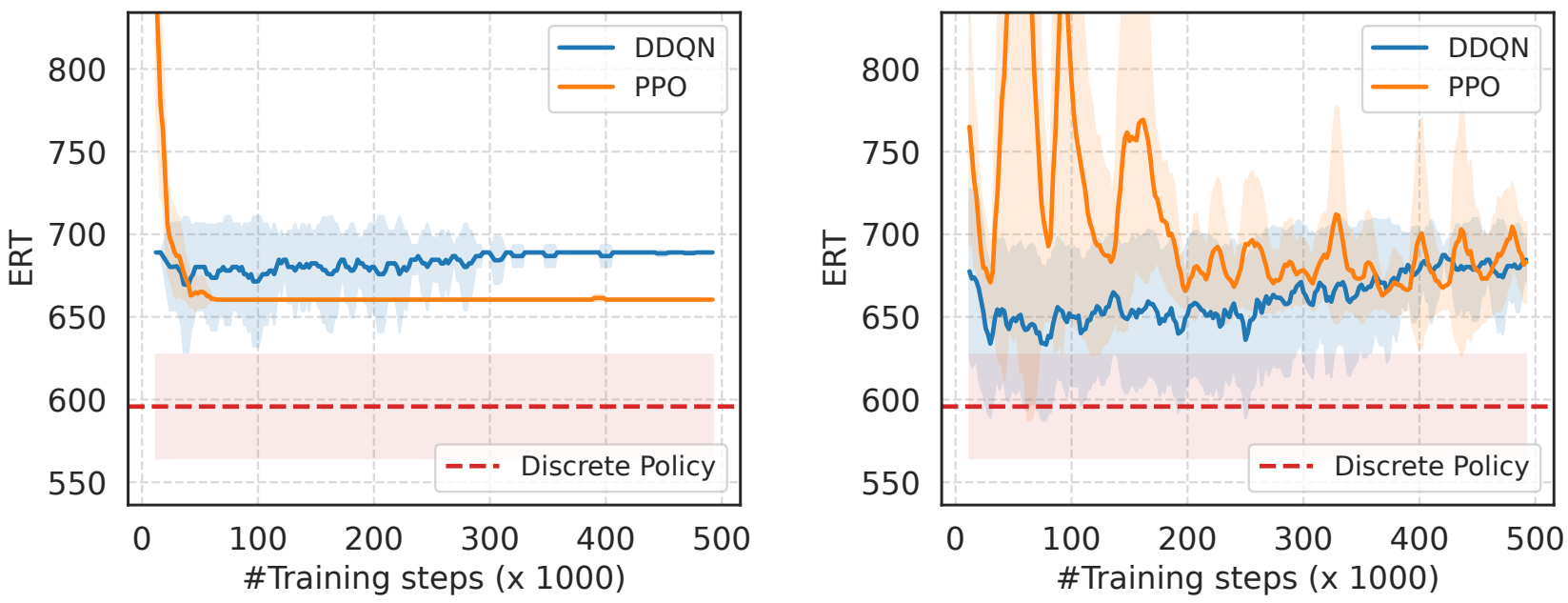

Fig. 3. Learning curves of DDQN and PPO using naïve (left) and scaled (right) reward functions for the problem size of 100.

**Naïve Rewards Induce Learning Stagnation.** Figure 3 (left) shows an example of DDQN and PPO learning curves on $n = 100$, where we plot the agent's average ERT. DDQN initially converges to some effective policies but subsequently diverges and stagnates throughout the remainder of the training process, whereas PPO learning improves during the first 40,000 training steps, but entirely fails to identify a good policy before stagnating. Both DDQN and PPO exhibit consistent stagnation during the latter phase of training across all RL runs and all four problem sizes evaluated. This issue makes the RL agent incapable of exploiting knowledge of well-performing policies, likely leaving a user with a far-from-optimal policy at the end of the training process, especially when we have a limited budget and cannot afford thorough evaluations to select the best-performing policy.

**Reward Scaling.** We have shown two learning limitations that affect both DDQN and PPO on the OneMax-DAC benchmark. To overcome these challenges, we study the impact of *reward scaling* on deep-RL learning performance. Our study is inspired by the fact that the original reward function $r_t = \Delta f_t - E_t$ is significantly influenced by the selected action. More concretely, from Algorithm 1, we can infer that $E_t \approx 2\lambda$. To maximize the returns, the agent would be biased towards choosing actions that minimize $E_t$, resulting in a bias towards choosing smaller $\lambda$ values. In fact, we see that the RL agents get stuck consistently in policies that dominantly select $\lambda = 1$ (the smallest $\lambda$ in our action space) across the entire state space. There is a need to limit the reward range, where the normalization[1] mechanisms are proved as simple yet effective solutions [Tessari and Iacca 2022; Xu et al. 2024]. The normalization process enhances the deep-RL's ability to discover the appropriate parameters $\theta$ much more efficiently [Sullivan et al. 2024; Van Hasselt et al. 2016b]. More specifically, to mitigate the variance scaling of $\lambda$ in the portfolio as the problem size increases, we scale the original reward function by the problem size: $r_t = \left(\Delta f_t - E_t\right)/n$. The scaled reward value thus ranges from $\frac{-2\max(\lambda)}{n}$ to $\frac{(n-1)}{n}$.

Table 1. ERT (± std) normalized by problem size $n$ among theory-derived policies (top), deep-RL approaches with two designed reward functions (middle), and the optimal policy (bottom) across four problem dimensions. Blue: RL outperforms theory-based. Bold: best non-optimal ERT.

| | | ERT(↓) | | | |
|---|---|---|---|---|---|
| | | $n = 50$ | $n = 100$ | $n = 200$ | $n = 300$ |
| $\pi_{\texttt{cont}}$ | | 5.448(1.53) | 5.826(1.18) | 6.167(0.97) | 6.278(0.84) |
| $\pi_{\texttt{disc}}$ | | 5.480(1.49) | 5.934(1.28) | 6.244(0.98) | 6.298(0.84) |
| *Deep-RL* | *Reward* | | | | |
| DDQN | Naïve | 5.428(1.86) | 6.437(1.94) | 7.045(1.53) | 8.615(2.16) |
| DDQN | Scaled | **5.105(1.56)** | 5.836(1.40) | 6.822(1.58) | 6.165(0.88) |
| PPO | Naïve | 5.534(1.99) | 6.670(1.97) | 7.806(2.09) | 8.531(2.12) |
| PPO | Scaled | 5.526(1.94) | 6.401(1.69) | 7.406(1.63) | 8.041(1.73) |
| $\pi_{\texttt{opt}}$ | | 4.928(1.42) | 5.313(1.10) | 5.609(0.89) | 5.750(0.75) |

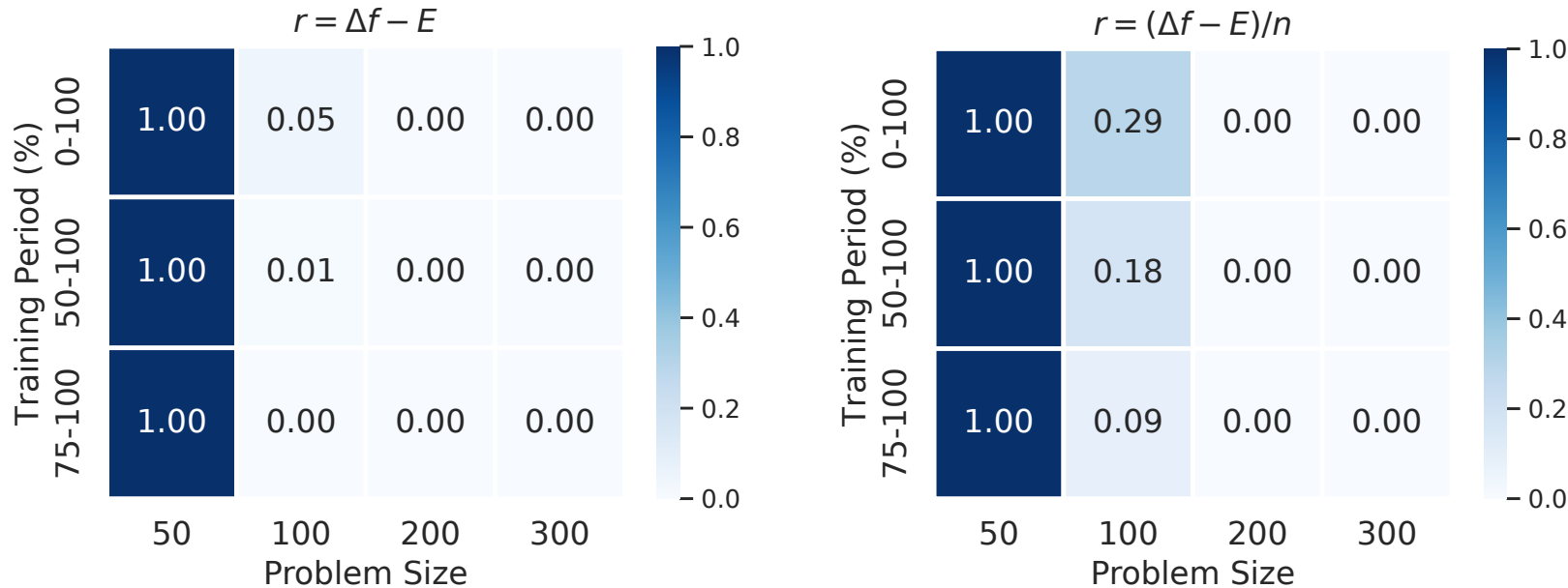

Fig. 4. DDQN performance (as HR across different training periods) using naïve (left) and scaled (right) reward functions.

Table 1 once again highlights the observation that the original reward function used by DDQN performs well in the relatively small problem size of 50, while PPO struggles to achieve similar results. DDQN with original reward function, however, becomes ineffective for the larger problem sizes. A similar observation is also made in the reward scaling, which are very promising for the two smaller problem sizes, $n \in \{50, 100\}$, but gradually becomes poor for larger problem sizes. On the contrary, PPO exhibits poor performance in both original and scaled reward functions, and this issue persists consistently across all problem sizes. Optimistically, normalizing the reward function can slightly enhance the learning of PPO, but the improvement is negligible compared to DDQN.

[1] We have also experimented with reward scaling, as proposed in [Engstrom et al. 2019] for policy-based deep-RL, where reward values are normalized by the standard deviation of the cumulative rewards obtained during rollouts. However, this method did not yield any noticeable improvements.

Examining the box plot in Figure 2 again, particularly focusing on the orange and red boxes, reveals that the scaling mechanism can provide a slight improvement in cases where $n \in \{200, 300\}$. By comparing the gap percentage to the discrete theory, the reward scaling function demonstrates exceptional performance compared to the conventional reward function across both approaches of deep-RL. However, Figure 3 (right) reveals their ERT diverges significantly and the best expected runtimes reported in the Table 1 for the reward scaling come from the early phases of learning.

Of the two deep-RL algorithms examined, only DDQN achieves performance comparable to that of a discrete policy. We therefore use the heatmap to present an overview of the HR across 10 DDQN runs. We analyze the learned policies across three distinct phases of the training process: the entire duration (0%-100%), the latter half (50%-100%), and the last quarter (75%-100%). By segmenting these periods, the heatmap helps identify potential divergence points. In an ideal scenario, where the RL agent progressively learns from the environment, we expect a consistent growth in the HR across all three phases. As shown in Figure 4, the smallest problem size achieves HR of 1 for all three periods, as the difference between good and poor policies in this setting is minimal. We thus need to analyze larger problems to gain a clearer understanding of the landscape. Generally, reward scaling effectively addresses scalability, particularly in the setting of problem size $n = 100$, where the agent discovers several good policies at the outset, achieving $\text{HR}@(0\% - 100\%) = 0.29$ compared to 0.05 with the original reward function. However, these HRs begin to decline over time, indicating the occurrence of divergence in both the original and scaled reward functions. More concretely, the heatmap for $n = 100$ under the scaled reward function reveals that HRs decrease over the three analyzed periods: $\text{HR}@(0\% - 100\%) = 0.29$, $\text{HR}@(50\% - 100\%) = 0.18$, and $\text{HR}@(75\% - 100\%) = 0.09$. A similar observation is evident in the heatmap for the conventional reward signal. For problems of size $n = 200$ and $n = 300$, the divergence issue becomes more severe, as HR = 0 observed in all periods.

## 4 Deep-RL Failures in OneMax-DAC

The experiments in the previous section revealed that both naïve and scaled reward functions suffer from learning stagnation, where deep-RL agents initially show learning improvement but subsequently experience stagnation for the remainder of the training. This counterintuitive behavior, where additional training leads to worse performance, that suggests fundamental issues with exploration in the OneMax-DAC environment. In this section, we diagnose the root causes of these failures and demonstrate why standard exploration mechanism are insufficient.

We define *stagnation* as the phenomenon where trained RL agents subsequently fail to maintain or improve their policies, leading to performance degradation with continued training. This behaviour manifests when the RL agent becomes trapped in sub-optimal policies and is unable to escape.

To understand how policies evolve during training, we analyze policy changes using pairwise differences between consecutive evaluations: $D(\pi_t, \pi_{t-1}) = \sum_{s \in \mathcal{S}} \mathbb{1}[\pi_t(s) \neq \pi_{t-1}(s)]$, where $\mathcal{S}$ is the full set of environment states, while $t$ and $\pi$ are the current evaluation iteration and the corresponding policy, respectively. Higher values indicate greater policy variation, suggesting active exploration, while low values suggest convergence or stagnation. In well-functioning RL training, we expect substantial policy changes initially (exploration phase) followed by gradual stabilization (convergence phase).

### 4.1 DDQN Analysis

DDQN learning with the original reward function (blue line) in Figure 5 (left), however, lacks changes in the beginning: its pairwise difference is always stably close to zero. This explains why, in Figure 3, the agent struggles to decrease runtime and finally stagnates. Learning with the scaled reward function (orange line) in Figure 5 (left), on the other hand, follows our expected pattern, which is why it performs better in Figure 3. Nevertheless, the fact that it still cannot overcome the stagnation challenge suggests that the agent does not explore the environment enough. We thus look for an effective solution to encourage exploration during learning.

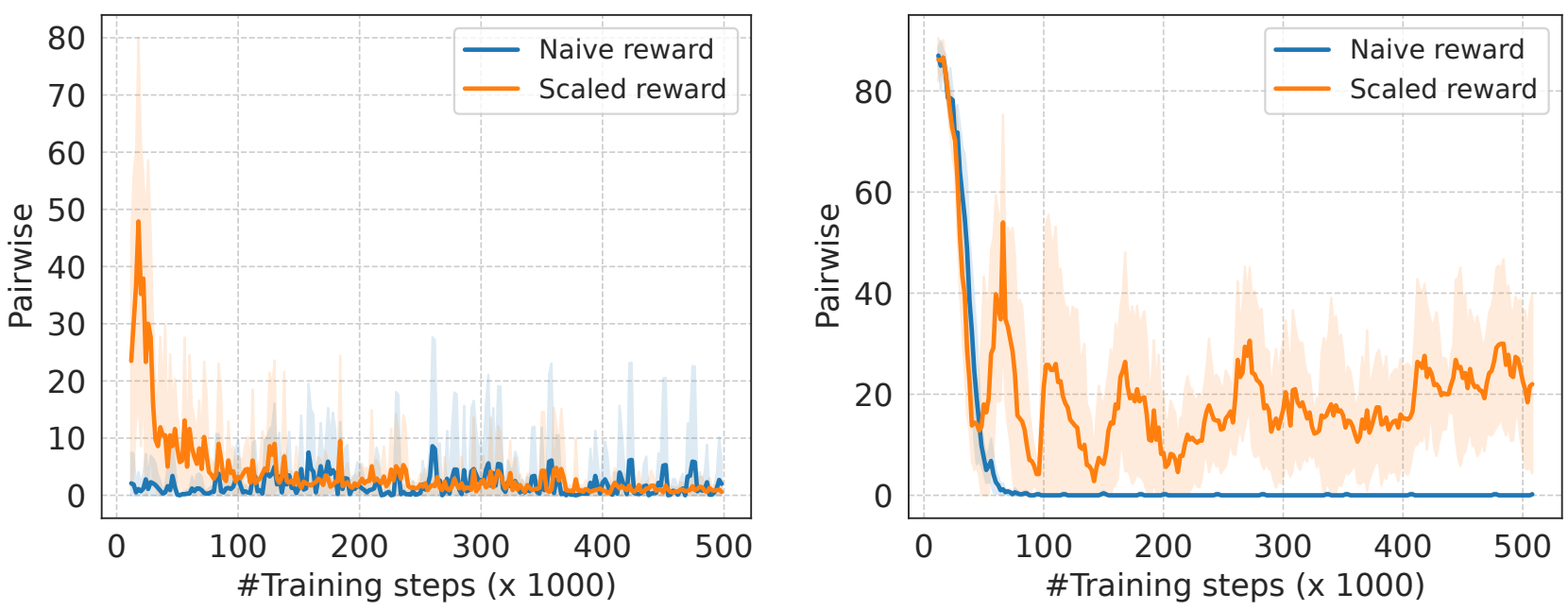

Fig. 5. Examples of the pairwise difference of DDQN (left) and PPO (right) for a problem size of 100.

**Random Exploration with $\epsilon$-Greedy.** In the literature, a popular choice to encourage exploration in deep $Q$-learning is $\epsilon$-greedy [Sutton 1988]. We tried various values of $\epsilon \in \{0.2, 0.3, 0.4, 0.5\}$ for the problem $n = 100$, using the DDQN setting described in Section 3. The top row in Figure 6 shows that the evaluated ERT still diverges until the end of the training budget, despite some slight improvements over the default setting ($\epsilon = 0.2$). For instance, learning with the original reward function achieves the best HR when $\epsilon = 0.3$, still the HR never exceeds 0.1 (middle row in Figure 6). We believe that the simple $\epsilon$-greedy strategy, which only explores the action space uniformly randomly, does not effectively improve exploration, as we can see in the bottom row in Figure 6 that the pairwise difference curves of different $\epsilon$ values are barely indistinguishable. This motivates our next study in Section 5 on employing a more sophisticated mechanism for improving exploration in value-based deep-RL, namely *reward shifting* [Sun et al. 2022].

### 4.2 PPO Analysis

PPO exhibits distinct issues depending on the reward function used, highlighting the algorithm's sensitivity to reward design in this environment. With the naïve reward function, PPO demonstrates premature convergence to suboptimal policies (blue line in Figure 5 (right)). The pairwise difference analysis shows rapid decline in policy variation, indicating that the agent quickly settles on a fixed strategy before adequately exploring alternatives. This behavior aligns with PPO's design philosophy of maintaining policy stability, but proves counterproductive when initial policies are far from the optimal (theory-derived policy in this case). Conversely, PPO with scaled rewards (orange line in Figure 5 (right)) maintains high policy variation throughout training but struggles to converge within a reasonable training budget. This pattern suggests that reward scaling provides sufficient exploration incentive but at the cost of learning stability, preventing the agent from consolidating knowledge about effective policies.

**Entropy Regularization.** In policy-based RL, one common approach to preventing premature convergence and encouraging exploration is to augment the objective with an entropy term. This idea is broadly referred to as *entropy regularization*; when the entropy term is treated as a central principle rather than a heuristic, it is known as the maximum entropy RL framework [Ahmed et al. 2019; Haarnoja et al. 2017, 2018a]. Theoretically, an additional intrinsic reward is incorporated into

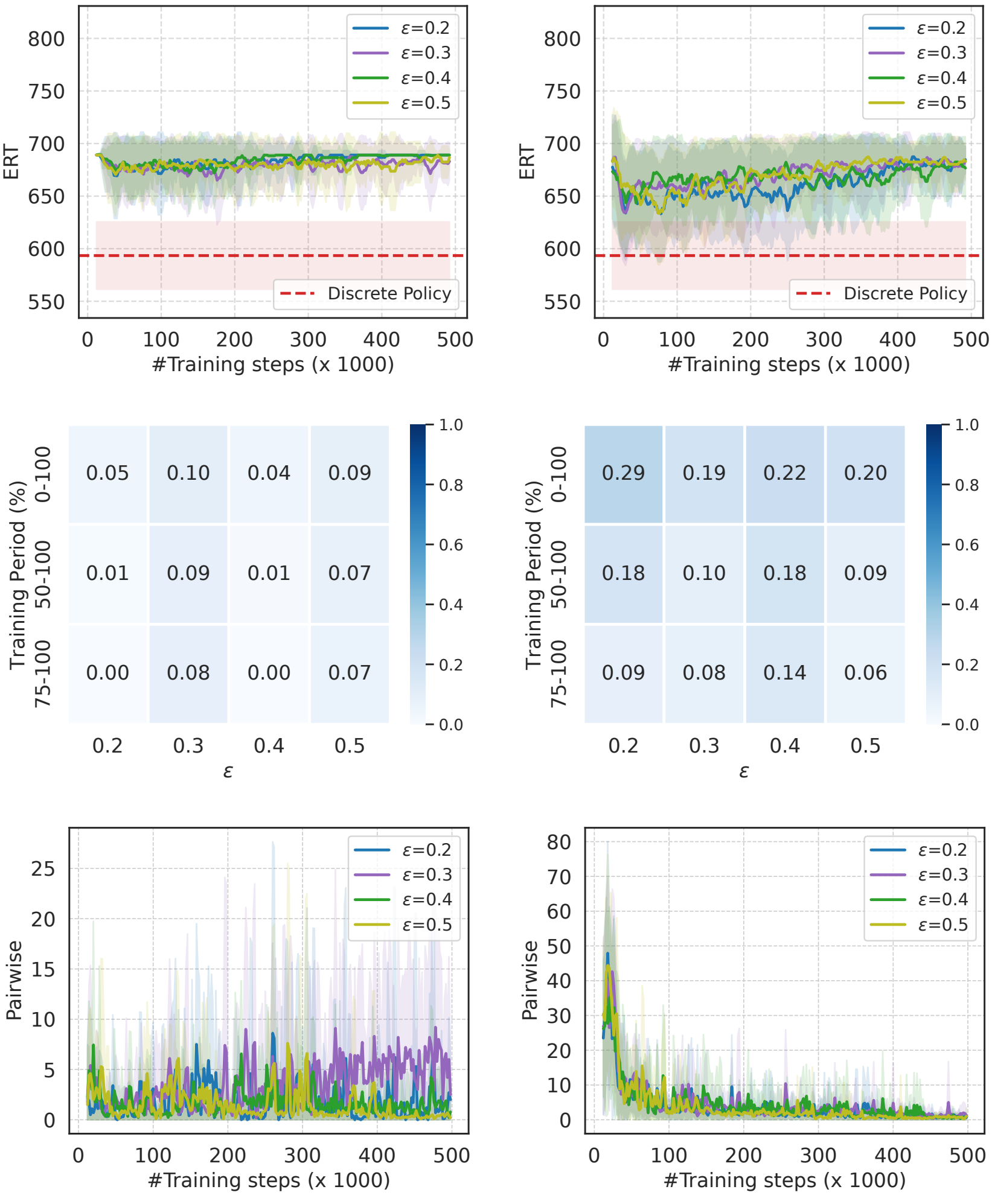

Fig. 6. $\epsilon$-greedy with various $\epsilon$ values applied on DDQN with naïve (left) and scaled (right) reward functions in a problem size of $n = 100$. The top row shows ERTs, the middle row presents HRs, and the bottom row displays pairwise differences.

the original objective, encouraging the policy to maximize entropy at each visited state:

$$\pi^* = \underset{\pi}{\operatorname{argmax}}\ \mathbb{E}_{\tau \sim \pi}\Big[\sum_{t=0}^{\infty} \gamma^t \big(r(s_t, a_t, s_{t+1}) + \beta \mathcal{H}(\pi(\cdot|s_t))\big)\Big], \tag{3}$$

where the objective is to find an optimal policy $\pi^*$ that maximizes the expected sum of future rewards while simultaneously maximizing its own entropy. Here, a policy $\pi(a_t|s_t)$ defines the agent's behavior by specifying the probability of taking action $a_t$ in a given state $s_t$. A trajectory $\tau = (s_0, a_0, s_1, a_1, \cdots)$ is a sequence of states and actions generated as the agent follows its policy within the environment. The expectation $\mathbb{E}_{\tau \sim \pi}$ is thus calculated over the distribution of these trajectories. In this equation, $\mathcal{H}(\pi(\cdot|s_t))$ is the entropy of the policy at state $s_t$, and $\beta$ is a coefficient that controls the trade-off between reward and entropy.

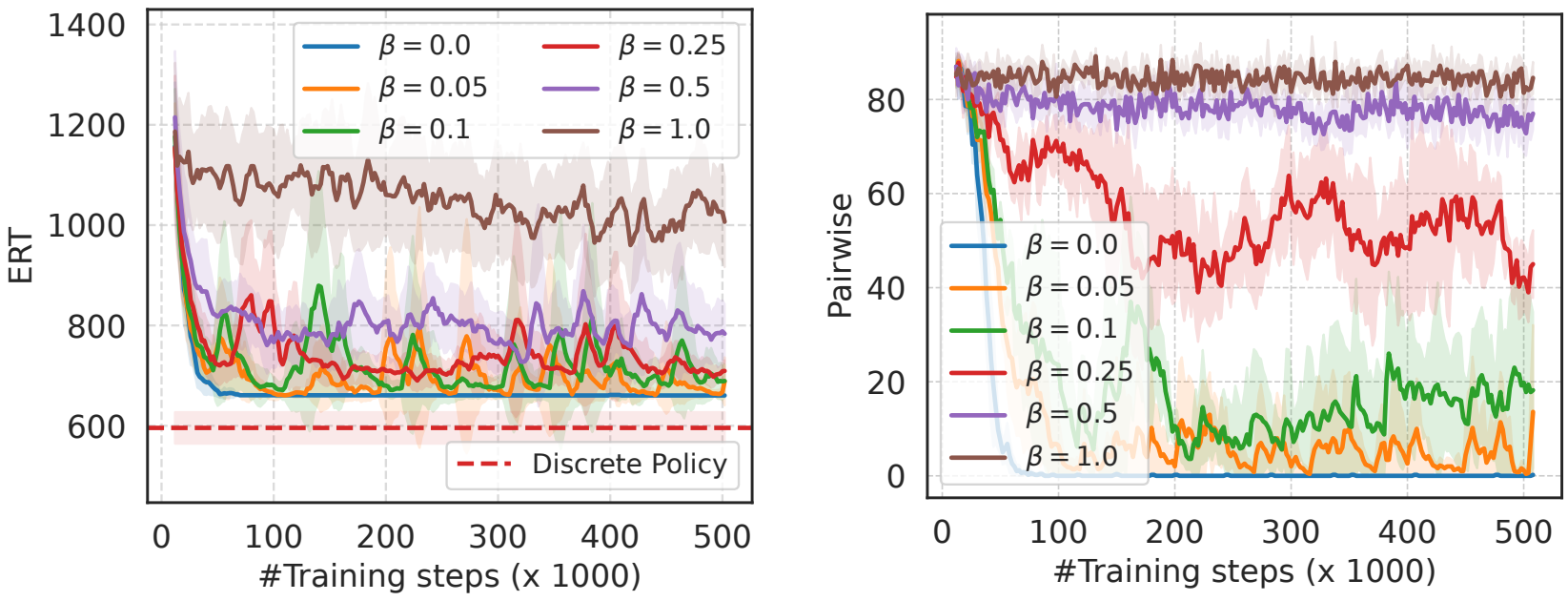

Fig. 7. Learning curves (left) and pairwise differences (right) across six different entropy coefficients of PPO, using the naïve reward function in a problem size of 100.

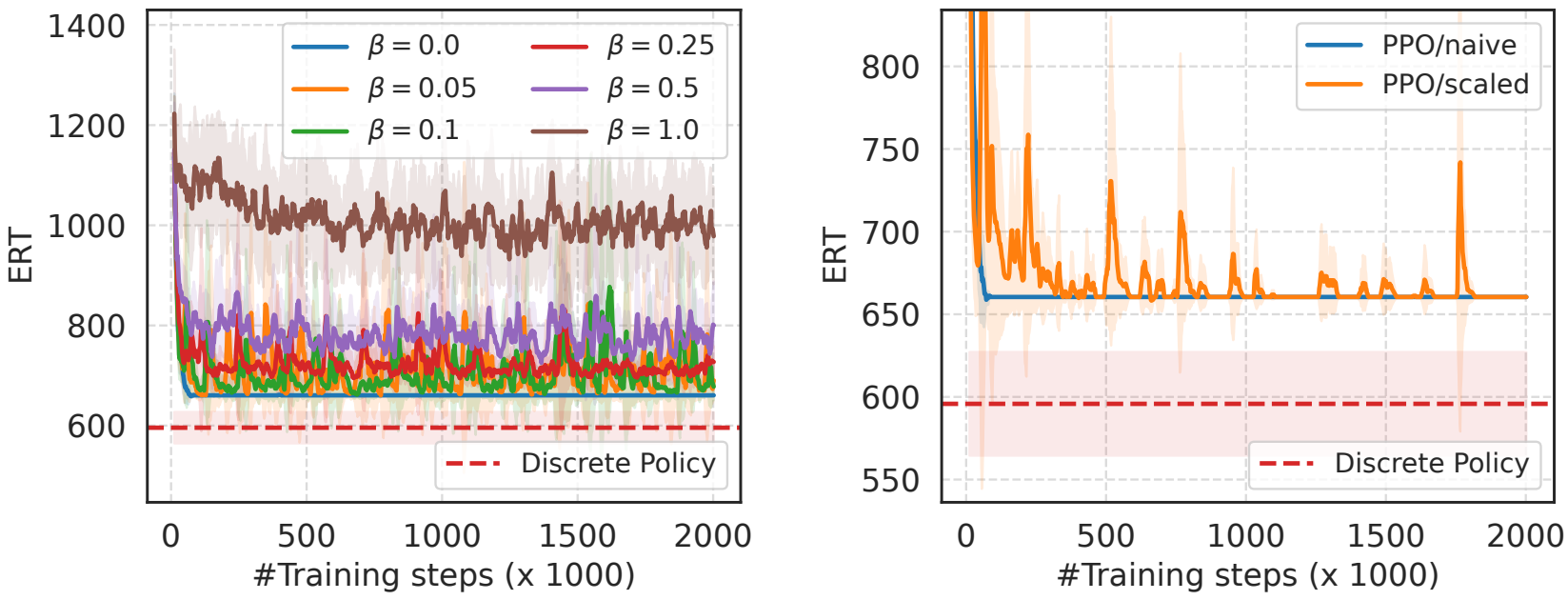

Fig. 8. Learning curves of PPO with an extended training budget of 2 million steps for a problem size of 100. Left: Performance with varying entropy coefficient values. Right: Comparison of PPO using naïve and scaled reward functions.

We investigate the effect of entropy regularization on the performance of the naïve reward function by varying the value of $\beta$ in the range $\{0.0, 0.05, 0.1, 0.25, 0.5, 1.0\}$ for $n = 100$. We use the PPO setting described in Section 3, where the default value of $\beta = 0.0$ indicates that entropy regularization is not applied. Figure 7 (left) illustrates that the entropy regularization encourages the PPO agent to explore the environment rather than getting stuck with a suboptimal policy. Figure 7 (right) shows that the higher the value of the entropy coefficient, the more diverse policies are discovered and evaluated. In particular, the smallest value of $\beta = 0.05$ yields promising results, where it is sufficient to increase exploration while ensuring that the learning process remains under control. However, the learning curves highlight that the training process is unstable and cannot converge to an optimal policy within the default training budget, which is similar to the observation of the performance using a scaled reward function.

**Extended Training Budget.** The slow convergence observed with naïve reward function (supported by entropy regularization) and scaled reward function raises the question of whether PPO simply requires more training time to stabilize. We extend the training budget up to 2 million steps. Figure 8 demonstrates two distinct problem patterns. For the entropy regularization with naïve reward function, the higher entropy coefficients lead to persistent fluctuations without convergence (Figure 8, left). The learning curves show that while entropy regularization prevents premature convergence to the suboptimal policy, it fails to provide sufficient guidance for discovering and stabilizing on better policies. On the other hand, PPO with scaled rewards initially shows promise but gradually deteriorates (Figure 8, right), eventually getting stuck at the suboptimal policy as the naïve reward function. The extended training demonstrates that the challenges faced by PPO are not merely due to insufficient training time but reflect fundamental algorithmic limitations within this environment.

## 5 Addressing Under-Exploration through Reward Shifting in DDQN

As detailed in Section 4.1, the under-exploration issue observed in this benchmark necessitates a more principled approach for resolution. In the literature, several studies [Bellemare et al. 2016; Choshen et al. 2018; Osband et al. 2016; Strehl and Littman 2004] have revealed that relying solely on the $\epsilon$-greedy strategy is inadequate to address the issue of under-exploration. These studies suggest more systematic approaches that encourage agents to prioritize visiting specific states, thereby improving exploration. Reward shaping [Laud 2004; Ng et al. 1999; Randløv and Alstrøm 1998] is a robust strategy in addressing the exploration challenge in the value-based RL training [Dey et al. 2024; Forbes et al. 2024; Ma et al. 2024; Sun et al. 2022]. The idea is to incorporate an external reward factor, $\mathcal{R}' = \mathcal{R} + F$ where $F$ is a shaping function, into the original reward $\mathcal{R}$. The external factor not only assists in stabilizing the training process but also accelerates the convergence of the RL algorithm. The stabilization arises from a sufficient trade-off between exploitation and exploration. Reward shaping can accelerate training because the agent needs to minimize the number of steps within an episode [Burda et al. 2018]. If the agent takes unnecessary steps, the cumulative reward becomes more negative, reducing the overall reward.

The effectiveness of reward shaping in addressing the hard exploration challenge lies in the fact that the $Q$-network is trained under the assumption of *optimistic initialization* [Brafman and Tennenholtz 2002; Even-Dar and Mansour 2001; Szita and Lőrincz 2008]. The values of the $Q$-network are initialized with optimistically high values: $\tilde{Q}_0(s, a; \theta) \geq \max_{a'} Q^*(s, a')\ \forall(s, a)$, where $\tilde{Q}_0(\cdot)$ represents the $Q$-values of the approximate network at the initial step and $Q^*(\cdot)$ is the true optimal $Q$-values. During the training, the trained $Q$-network gradually relaxes to estimates of the expected returns based on the observations.

Inspired by the theory of reward shaping, in [Sun et al. 2022] *reward shifting* was introduced ($\mathcal{R}' = \mathcal{R} + b$ with $b \in \mathbb{R}$) to encourage the agent to explore its environment and escape suboptimal points in the optimization landscape. In the context of reward shifting, it is not necessary to directly initialize the network's parameters to obtain higher $Q$-values. Instead, the $Q$-values of the optimal policy $\pi^*$ are shifted downward by a constant relative to the original situation, as shown in Figure 9(a). A proof of the relationship between external bias and the shifting distance is provided in the supplementary material. This shift assumes that all values associated with the approximate policy $\tilde{\pi}$ are initially higher than those of the true optimal policy $\pi^*$. In the network update, the $Q$-values of the chosen action at step $t$ is pulled closer to the shifted true optimal policy. Meanwhile, the $Q$-values of the unchosen actions are maintained at relatively high levels as shown in Figure 9(b). In this manner, the selected action at step $t + 1$ will less frequently adhere to the knowledge of the optimal policy, resulting in a more extensive exploration of the environment. In contrast, initializing with lower $Q$-values of $\tilde{\pi}$ or setting the optimal policy $\pi^*$ higher can steer the agent toward exploitation rather than exploration; this concept is commonly referred to as *pessimistic initialization.* By demonstrating some empirical results, Sun et al. [2022] conclude that an upward shift is associated with a positive value of $b^+$, which leads to conservative exploitation, while a downward shift corresponds to a negative value of $b^-$, inducing curiosity-driven exploration.

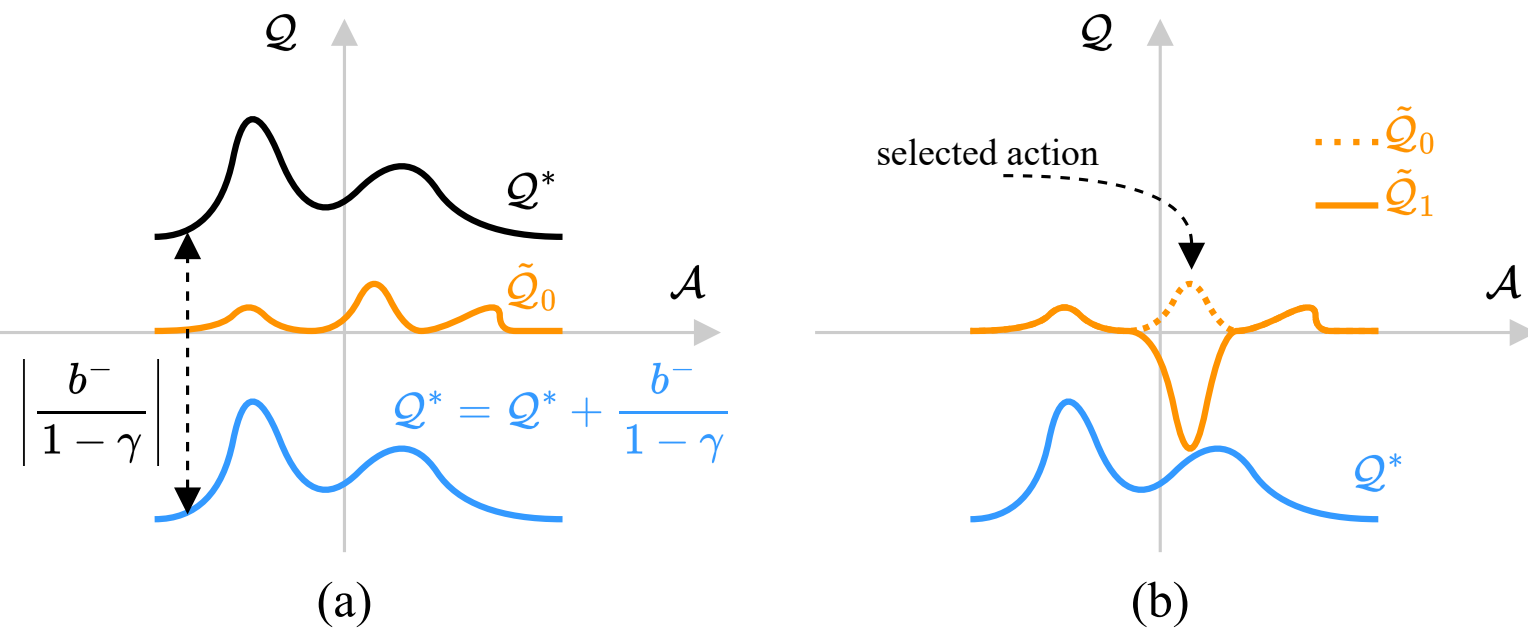

Fig. 9. Illustrative landscapes of $Q$-values at two specific steps: (a) The optimal $Q^*$ is uniformly shifted downward by a constant $\left|\frac{b^-}{1-\gamma}\right|$, without altering the overall landscape structure. (b) Upon selecting an action and visiting a specific state, its corresponding state-action value $\tilde{Q}$ is updated towards the optimal $Q^*$, while the $\tilde{Q}$-values of unselected actions remain comparatively higher.

Table 2. ERT (± std) normalized by the problem size $n$ of optimal, theory-derived and RL-based DAC policies with a fixed shifting bias and with adaptive shifting $b_a^-$. Blue: RL outperforms theory-based. Bold: best non-optimal ERT. Underline: not significantly worse than the best (with a confidence level of 95%).

| | **ERT(↓)** | | | |
|---|---|---|---|---|
| | $n = 50$ | $n = 100$ | $n = 200$ | $n = 300$ |
| $\pi_{\texttt{cont}}$ | 5.448(1.53) | 5.826(1.18) | 6.167(0.97) | 6.278(0.84) |
| $\pi_{\texttt{disc}}$ | 5.480(1.49) | 5.934(1.28) | 6.244(0.98) | 6.298(0.84) |
| $r = \Delta f - E + 1$ | 5.658(2.01) | 6.810(2.14) | 7.893(2.04) | 8.603(2.16) |
| $r = \Delta f - E + 3$ | 5.658(2.01) | 6.810(2.14) | 7.893(2.04) | 8.613(2.16) |
| $r = \Delta f - E + 5$ | 5.649(1.99) | 6.786(2.11) | 7.891(2.05) | 8.546(2.14) |
| $r = \Delta f - E - 1$ | 4.994(1.43) | 5.681(1.33) | 6.576(1.27) | 7.042(1.23) |
| $r = \Delta f - E - 3$ | 5.049(1.41) | 5.421(1.15) | 5.975(1.01) | 6.481(1.01) |
| $r = \Delta f - E - 5$ | 5.002(1.42) | 5.520(1.16) | **5.671(0.92)** | 6.118(0.88) |
| $r = \Delta f - E + b_a^-$ | **4.919(1.44)** | 5.444(1.19) | 5.723(0.93) | **5.990(0.85)** |
| $r = (\Delta f - E)/n + b_a^-$ | 4.927(1.41) | **5.409(1.19)** | 5.723(0.93) | 6.103(0.93) |
| $\pi_{\texttt{opt}}$ | 4.928(1.42) | 5.313(1.10) | 5.609(0.89) | 5.750(0.75) |

### 5.1 Reward Shifting with a Fixed Bias

Following Sun et al. [2022], we implement the shifting mechanism by adding a constant bias into the original reward function in Equation (2): $r_t = \Delta f_t - E_t + b$. To determine the optimal shifting bias $b$, we replicate the experiments conducted in previous sections, employing both negative and positive biases. We systematically evaluate the impact of fixed bias values $\{\pm 1, \pm 3, \pm 5\}$.

Table 2 shows the average ERT across 10 RL runs for each approach, together with the baselines, where adding the positive fixed shifts $b$, which range from +1 to +5 results in always higher ERT (i.e., worse performance). All settings associated with the negative biases outperform the positive options. More explicitly, they are better than the discretized theory policy $\pi_{\texttt{disc}}$ and even outperform the original theory policy $\pi_{\texttt{cont}}$ in several cases. This observation consolidates our conjecture about the under-exploration problem in using the original reward function, and that the RL agent should focus more on exploration than exploitation.

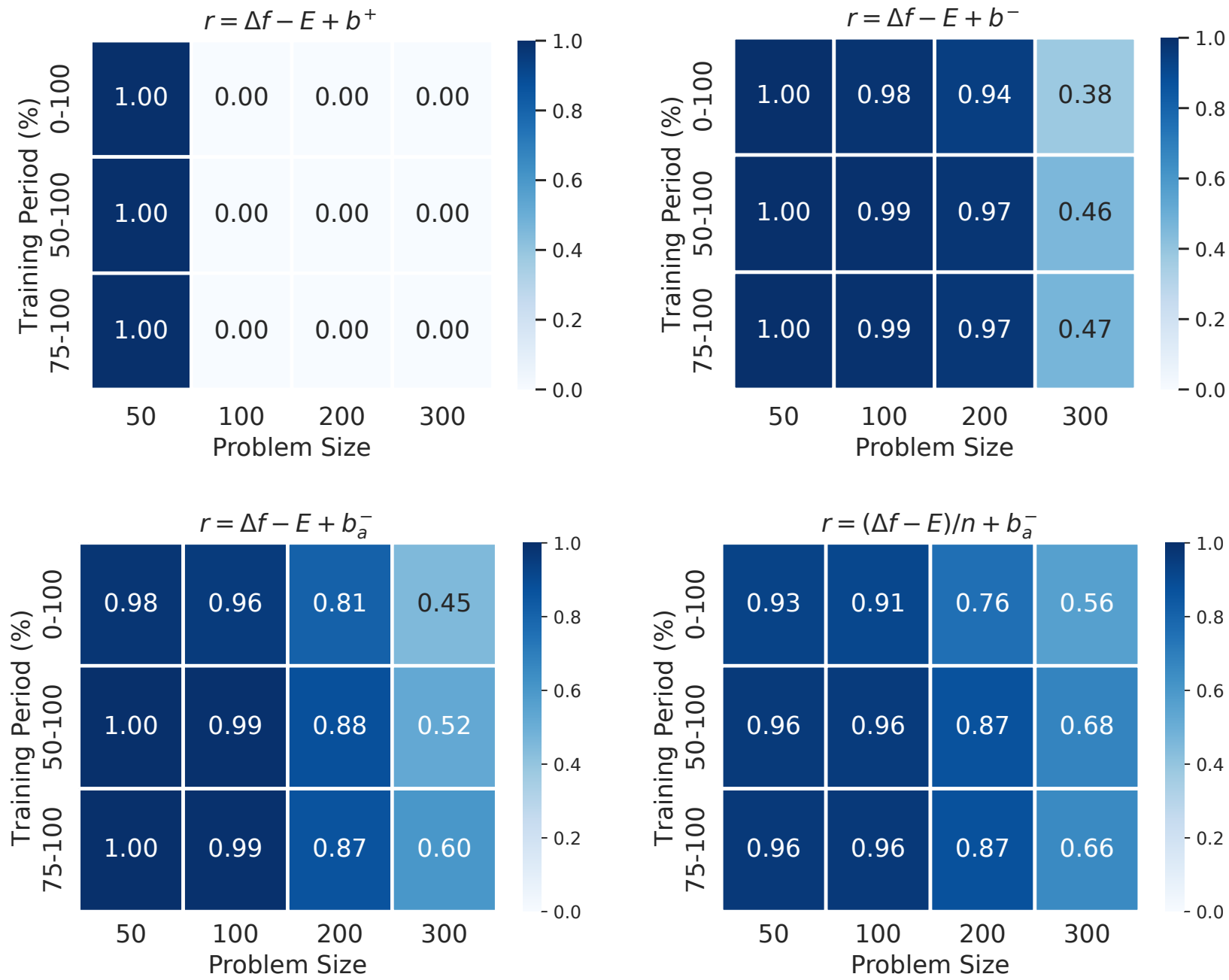

Fig. 10. DDQN performance (HR across training periods) using different reward functions: original reward function with fixed positive shifts (top left), fixed negative shifts (top right), adaptive shifted biases (see Section 5.2) applied to the original reward (bottom left) and the scaled reward (bottom right). For the fixed negative shift condition, HRs are shown for the best-performing bias for each problem size.

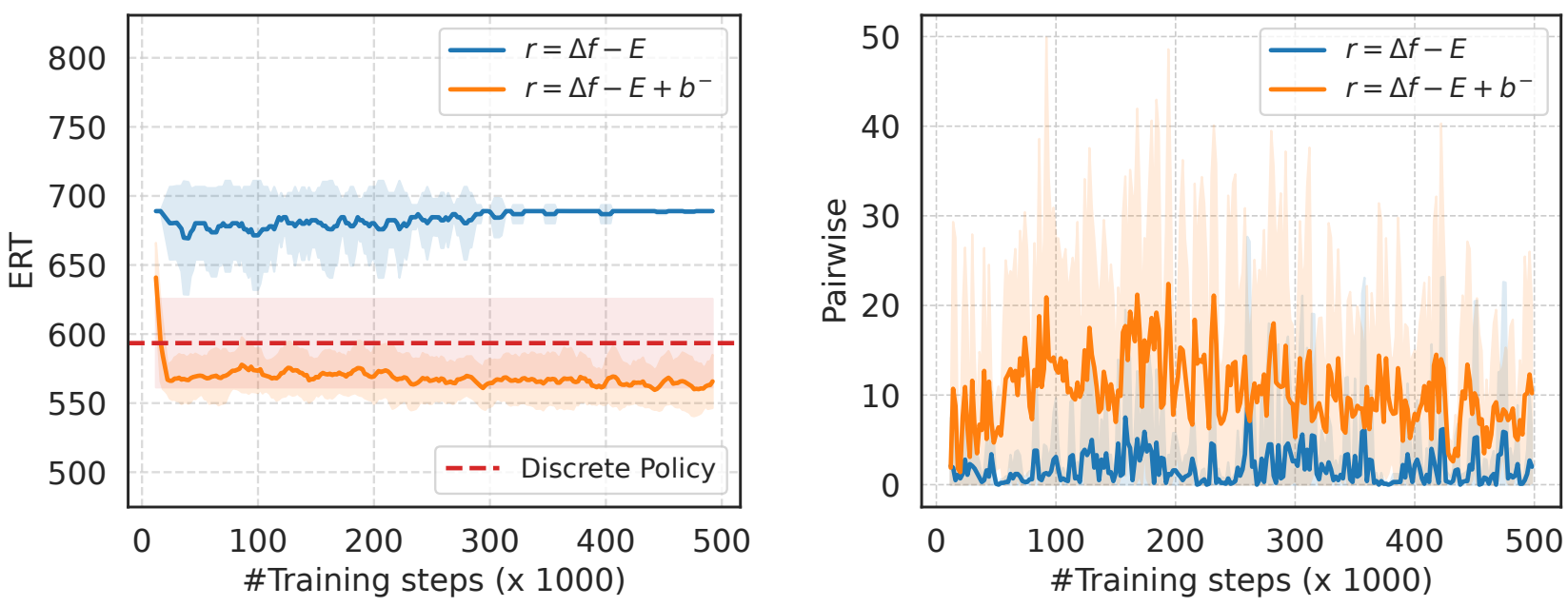

Fig. 11. Effectiveness of reward shifting for a problem size of 100 with $b^- = -3$: (left) ERT and (right) pairwise difference.

Figure 10 provides more details about the ability of each approach in terms of converging to a good policy, where positive values of the shift completely fail the task with a problem size larger than 50. In contrast, the robustness of the negative shiftings is demonstrated in the three problem sizes ranging from 50 to 200, where almost over 90% of the evaluated points adhere to the theory policy. Although this strength is not maintained when the problem size increases to 300, where the HR decreases by half, the negative reward shifting mechanism remains a promising solution to the divergence and stagnation problem as the HR does increase during the later part of the training process.

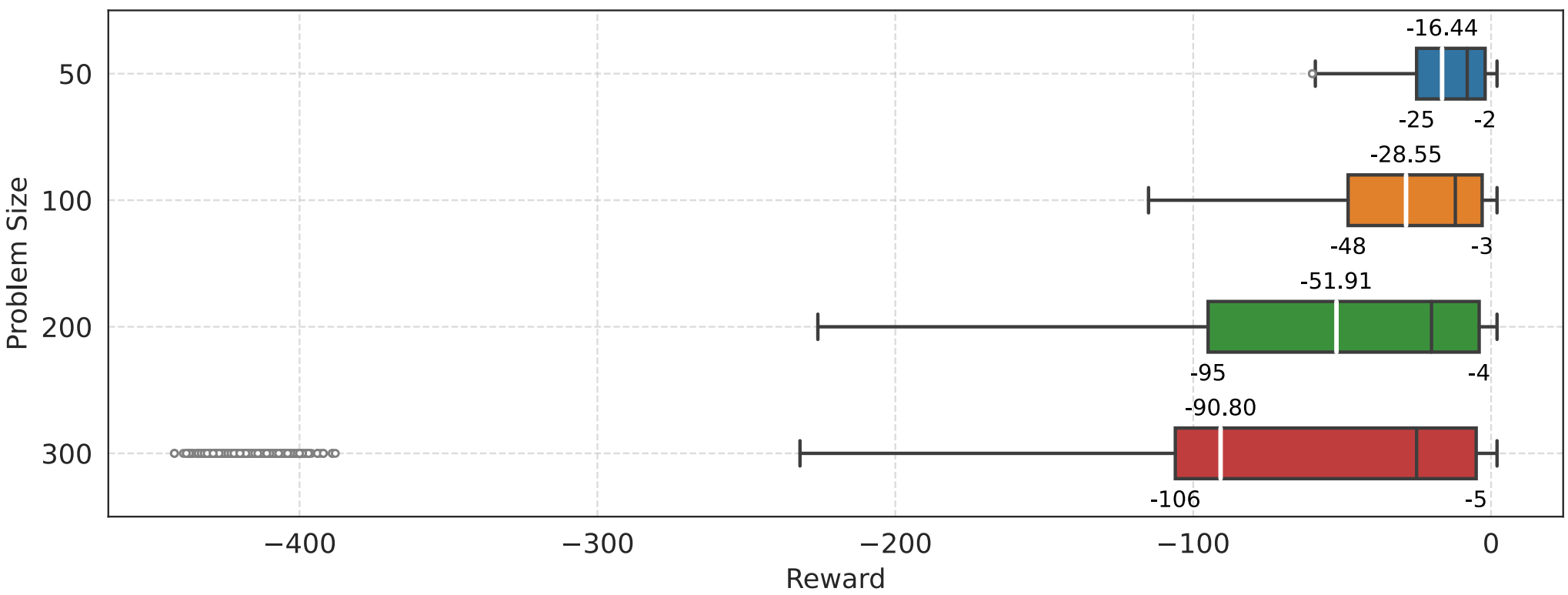

Fig. 12. Reward value distributions across problem sizes $n \in \{50, 100, 200, 300\}$. The quartile values $(Q_1, Q_3)$ are indicated by the borders of each box and their values are displayed below the boxes, while the mean $(\bar{R})$ is represented by a white stripe and its value is positioned above each box.

To assess the effectiveness of reward shifting in helping exploration, we analyze the policy changes when employing fixed shifting bias compared to the original reward function. Figure 11 clearly demonstrates the significant increase in policy changes when compared to those without shifting bias. The learned agent selects a variety of actions and gradually converges towards the desired outcome throughout the training process. Meanwhile, the agent walks in the correct direction toward the theory baseline as shown in the ERT plot.

## 5.2 Adaptive Shifting

While fixed bias values demonstrate effectiveness, determining optimal bias values for different problem sizes presents a practical challenge. As illustrated in Table 2, the best bias varies across problem sizes. Manual bias selection for each problem size is impractical and contradicts the goal of dynamic algorithm configuration. Sun et al. [2022] suggests using a *meta-learner* to control the value of $b$, but this would significantly increase the cost of learning. To address this limitation, we develop an *adaptive reward shifting* mechanism that automatically estimates the bias, thereby eliminating the need for parameter tuning per problem size. We examine reward values across various problem sizes, specifically $n \in \{50, 100, 200, 300\}$. For each problem size, we present the corresponding reward distribution throughout the 10,000 warm-up steps (refer to Figure 12). We identify two principal empirical patterns: (1) the reward range expands exponentially as problem dimensions increase; and (2) the best negative shifting bias in Table 2 is directly proportional to both the average reward and the total reward range. We therefore derive the following formula for deciding the adaptive shifting bias $b_{\text{a}}^{-}$:

$$b_{\text{a}}^{-} = 0.0052\, \bar{R}\, \frac{Q_1}{Q_3} \tag{4}$$

where $\bar{R}, Q_1, Q_3$ represent the average reward, first quartile, and third quartile, respectively. The multiplication factor of 0.0052 is determined through a few trials. Notably, this formula requires tuning the multiplication factor just once, regardless of the problem size.

The lower part of Table 2 demonstrates that our adaptive approach ($r_t = \Delta f_t - E_t + b_{\text{a}}^-$) achieves competitive performance compared to the best fixed bias configurations, consistently outperforming both theory-derived baselines across all tested problem sizes without requiring problem size-specific hyperparameter adjustment. A comprehensive comparison between the adaptive shifting proposed in [Nguyen et al. 2025a] and our new adaptive technique intuition is provided in the supplementary material, demonstrating the consistent improvement in learning stability and scalability of our approach.

### 5.3 Reward Shifting and Scaling

In Table 1, we have observed some potential of the reward scaling technique to improve learning performance. It motivates our study in this section where we investigate the combined effect of reward shifting and reward scaling. We therefore directly incorporate the adaptive shifting mechanism into reward scaling:

$$r_t = (\Delta f_t - E_t)/n + b_{\text{a}}^-. \tag{5}$$

As shown in Table 2, compared to the shifting-only reward function, the overall average ERT of the shifting-scaling version is marginally worse. However, there are two exceptions: when $n \in \{50, 100, 200\}$, the shifting-scaling version performs competitively to the adaptive reward shifting alone. Unfortunately, as the problem size increased, this combination began to reveal its disadvantages. It could no longer maintain the safe distance from the discrete theory-derived policies that adaptive reward shifting had maintained. Nevertheless, the hitting rates depicted in Figure 10 (bottom) expose the opposite behavior. In this case, the learning curves for shifting-scaling exhibit less stability compared to the reward shifting across three small-medium problem sizes. However, there is a slight improvement in the problem size of $n = 300$. In general, we do not observe a clear advantage in combining the two mechanisms.

## 6 Long-horizon Episodes in the OneMax-DAC Environment

In the previous section, we presented an effective solution for improving and stabilizing the learning of DDQN via reward shifting. In this section, we investigate another aspect of the learning: the challenge of dealing with long-horizon episodes. Based on the investigation, we look into the impact of the discount factor on the learning performance of both DDQN and PPO.

### 6.1 Problem of Planning Horizon in the OneMax-DAC

To quantify the relationship between the problem complexity and episode length, we evaluate 3 distinct policies across 1,000 trials: a random policy (randomly selecting population size), a supoptimal (or “all ones”) policy[2] and discrete theory-derived policy ($\pi_{\text{disc}}$). Figure 13 (left) clearly shows episode length distributions across problem sizes, with substantial variation between policies.

For the standard discount factor $\gamma = 0.99$, the effective planning horizon is $\mathbb{E}[T] = 100$ steps. Comparing this with observed episode lengths of $\pi_{\text{disc}}$ (green boxes in Figure 13, left), where the problem size of $n = 50$ requires $e^{4.2} \approx 67$ steps[3], while this number of $n = 100$ is $e^5 \approx 150$ steps (approximately 33% of episode rewards receive the discounting weight less than 0.5). Table 1 shows that DDQN performance (gap to $\pi_{\text{disc}}$) deteriorates as problem size increases: 0% gap for $n = 50$, 8% gap for $n = 100$. This degradation strongly correlates with the decreasing coverage of episodes by the planning horizon. We therefore assume that the exponential suppression of distant rewards explains why deep-RL agents perform competitively on smaller problems but fail dramatically on larger ones.

[2] We identify the suboptimal policy where the DDQN agent gets stuck, as shown in Figure 3, we observe that the DDQN agent consistently chooses $\lambda = 1$.

[3] The results in Figure 13 are presented on a logarithmic scale.

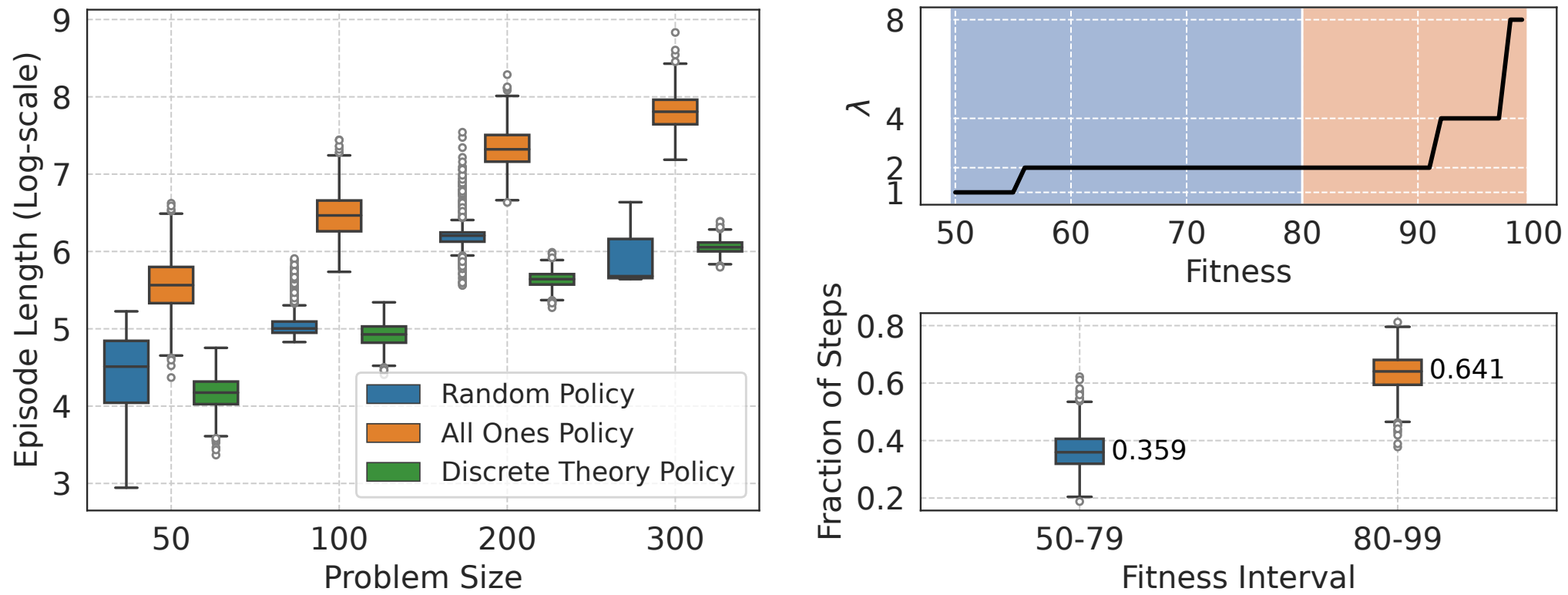

Fig. 13. Episode lengths for three distinct policies across four problem sizes (left); the shape of the discrete theory-derived policy for $f(x)/n \geq 0.5$ (top-right); and the distribution of number of steps (over episode length) per fitness interval for the problem size of 100 (bottom-right).

In addition, we examine how optimization steps are distributed across specific fitness intervals, as illustrated in Figure 13 ( right). Specifically, the total steps within the blue interval are represented by the blue box plot, while the total steps within the orange interval are shown in the orange box plot. We have already known that the orange interval (Figure 13, top-right) is the most informative because the policy rapidly adjusts the population size $\lambda$ within this interval to accelerate evolution, and the reward signals in this area are more meaningful than others. As shown in Figure 13 (bottom-right), over 64% of episode length concentrates in the later interval. If an agent fails to adequately plan its actions within this interval, it will be forced to make policy decisions based on incomplete or misleading rewards.

Based on these observations above, we suppose that deep-RL agent performance tends to drop when the effective planning horizon, $\mathbb{E}[T] = \frac{1}{1-\gamma}$, is shorter than the actual episode length. Rewards that fall outside this horizon are heavily discounted by $\gamma$, so the agent almost ignores them during training. We conjecture that performance improves when the discount factor is higher, since this extends the planning horizon, and the effect is usually stronger for larger problems where the gap between horizon and episode length is bigger. This understanding motivates our look into how the discount factor influences both learning performance and stability.

## 6.2 Discount Factor Analysis in Deep-RL

In standard MDPs, the discount factor serves two primary purposes: (1) ensuring bounded expected rewards in infinite-horizon problems, and (2) encouraging agents to prioritize short-term rewards for faster goal achievement [Amit et al. 2020; Sutton and Barto 1998]. However, setting the discount factor too low causes myopic behavior, while setting $\gamma = 1$ can lead to learning instability due to unbounded expected rewards [Jiang et al. 2015].

**Experiment Setup.** To validate our hypothesis in Section 6.1, we systematically evaluate discount factors $\gamma \in \{0.9, 0.99, 0.995, 0.9998, 1.0\}$ on both DDQN and PPO, while maintaining all other hyperparameters at default values (Section 3). For DDQN, we analyze all three designed reward functions, while PPO is examined with original and scaled reward functions only. We evaluate policy quality through ERT measurements and analyze learning convergence via learning curve, hitting rate (HR), and area under the curve (AUC) across small-to-medium problem sizes.

Table 3. ERT and AUC (± std) normalized by problem size $n$, over 1000 seeds of the best DDQN policies for reward functions {naïve, scaled, adaptive shifted} and discount factors $\gamma$, across $n \in \{50, 100, 200, 300\}$. Blue: RL outperforms theory-based. Bold: best non-optimal ERT/AUC. Underline: not statistically significantly different from the best ERT (Bonferroni-corrected paired two-sided t-tests with confidence level of 95%).

| | | $n = 50$ | | $n = 100$ | | $n = 200$ | | $n = 300$ | |
|---|---|---|---|---|---|---|---|---|---|
| | | ERT(↓) | AUC(↓) | ERT(↓) | AUC(↓) | ERT(↓) | AUC(↓) | ERT(↓) | AUC(↓) |
| $\pi_{\text{cont}}$ | | 5.448(1.53) | - | 5.826(1.18) | - | 6.167(0.97) | - | 6.278(0.84) | - |
| $\pi_{\text{disc}}$ | | 5.480(1.49) | - | 5.934(1.28) | - | 6.244(0.98) | - | 6.298(0.84) | - |
| *Reward* | $\gamma$ | | | | | | | | |
| Naïve | 0.9 | 5.611(2.08) | 2.319(0.06) | 6.708(2.04) | 4.393(0.03) | 7.721(1.83) | 5.911(0.02) | 8.210(1.74) | 6.528(0.03) |
| | 0.99 | 5.428(1.86) | 2.538(0.09) | 6.437(1.94) | 4.320(0.41) | 7.045(1.61) | 5.857(0.10) | 8.615(2.16) | 6.511(0.00) |
| | 0.995 | 5.010(1.51) | 0.708(0.40) | 5.856(1.48) | 2.098(0.29) | 6.720(1.39) | 5.268(0.09) | 7.201(1.35) | 6.121(0.08) |
| | 0.9998 | 4.934(1.45) | 0.203(0.10) | 5.505(1.24) | 0.924(0.25) | 5.875(0.95) | 3.533(0.47) | 6.132(0.88) | 4.456(0.74) |
| | 1.0 | 4.933(1.44) | 0.221(0.09) | 5.499(1.21) | 0.985(0.30) | 5.863(0.96) | 3.529(0.40) | 6.030(0.85) | 4.331(0.61) |
| Scaled | 0.9 | 5.585(2.03) | 2.320(0.04) | 6.706(2.08) | 4.393(0.03) | 7.494(1.75) | 5.917(0.02) | 7.916(1.58) | 6.534(0.03) |
| | 0.99 | 5.105(1.56) | 2.240(0.33) | 5.836(1.40) | 3.638(0.69) | 6.822(1.58) | 5.894(0.09) | 6.165(0.88) | 6.485(0.03) |
| | 0.995 | 5.007(1.51) | 1.003(0.19) | 5.762(1.38) | 1.826(0.40) | 6.322(1.21) | 5.139(0.14) | 6.799(1.21) | 6.207(0.16) |
| | 0.9998 | 4.937(1.42) | 0.721(0.13) | **5.397(1.19)** | 1.038(0.24) | 5.764(0.92) | 3.554(0.44) | 6.180(0.99) | 4.938(0.74) |
| | 1.0 | **4.914(1.42)** | 0.776(0.09) | 5.402(1.17) | 1.278(0.29) | **5.710(0.92)** | 3.598(0.58) | 6.097(0.97) | 5.133(0.58) |
| Adaptive Shifted | 0.9 | 5.564(2.04) | 2.336(0.04) | 6.718(2.12) | 4.394(0.03) | 7.120(1.56) | 5.909(0.02) | 7.263(1.36) | 6.312(0.11) |
| | 0.99 | 4.919(1.44) | **0.131**(0.03) | 5.444(1.19) | **0.343**(0.08) | 5.723(0.93) | **1.432**(0.56) | **5.990(0.85)** | **3.285**(1.18) |
| | 0.995 | 4.968(1.41) | 0.747(0.25) | 5.533(1.23) | 3.023(0.21) | 5.811(0.95) | 4.411(0.78) | 6.192(0.88) | 6.109(0.18) |
| | 0.9998 | 5.012(1.46) | 2.523(0.13) | 5.607(1.21) | 4.219(0.06) | 5.895(0.95) | 5.571(0.27) | 6.365(0.95) | 6.613(0.18) |
| | 1.0 | 5.037(1.52) | 2.643(0.19) | 5.630(1.23) | 4.263(0.07) | 5.903(0.92) | 5.609(0.19) | 6.283(0.93) | 6.677(0.17) |
| $\pi_{\text{opt}}$ | | 4.928(1.42) | - | 5.313(1.10) | - | 5.609(0.89) | - | 5.750(0.75) | - |

**DDQN Results.** Among the three reward functions and five discount factor values, the combination of the two highest values of $\gamma \in \{0.9998, 1.0\}$ with the scaled reward function emerges as the top-2 most robust settings in terms of ERT across 3 out of 4 tested problem sizes, as highlighted in bold in Table 3. On the other hand, the AUC metric indicates that an adaptive shifted reward function with the default discount factor (i.e., $\gamma = 0.99$) emerges as the most effective learning approach. This means its learning is more stable and capable of finding a good policy more quickly than other settings. These observations suggest that while larger discount factors alleviate long-horizon planning issues, they simultaneously lead to less stable learning dynamics, exposing a trade-off between effective horizon coverage and learning stability.

We now analyze the results of naïve and scaled reward functions (the upper two clusters in Table 3). Higher discount factors consistently improve performance across these two functions, especially from a problem size of $n = 100$ onwards, as highlighted in blue in Table 3. This observation confirms that planning horizon coverage is the limiting factor, as our hypothesis predicts. Although higher discount factors combined with the shifting reward function still partly outperform the baselines (as highlighted in blue in Table 3), there is a clear trend that a larger discount factor results in poorer ERT performance. The only optimistic observation is that the ERTs when equipping the shifting mechanism with the default value of gamma, are statistically indistinguishable[4] from the best settings in two problem sizes of $n \in \{50, 200\}$. While the ERT of adaptive reward shifting for $n = 100$ is slightly inferior to the others, even when a statistical test is applied, it is still considered acceptable as it surpasses the baselines. Notably, for the largest problem size ($n = 300$), the adaptive shifting mechanism proves advantageous in terms of runtime, completely outperforming both the naïve and scaled reward function settings.

[4]According to Bonferroni-corrected paired two-sided t-tests with a confidence level of 95%.

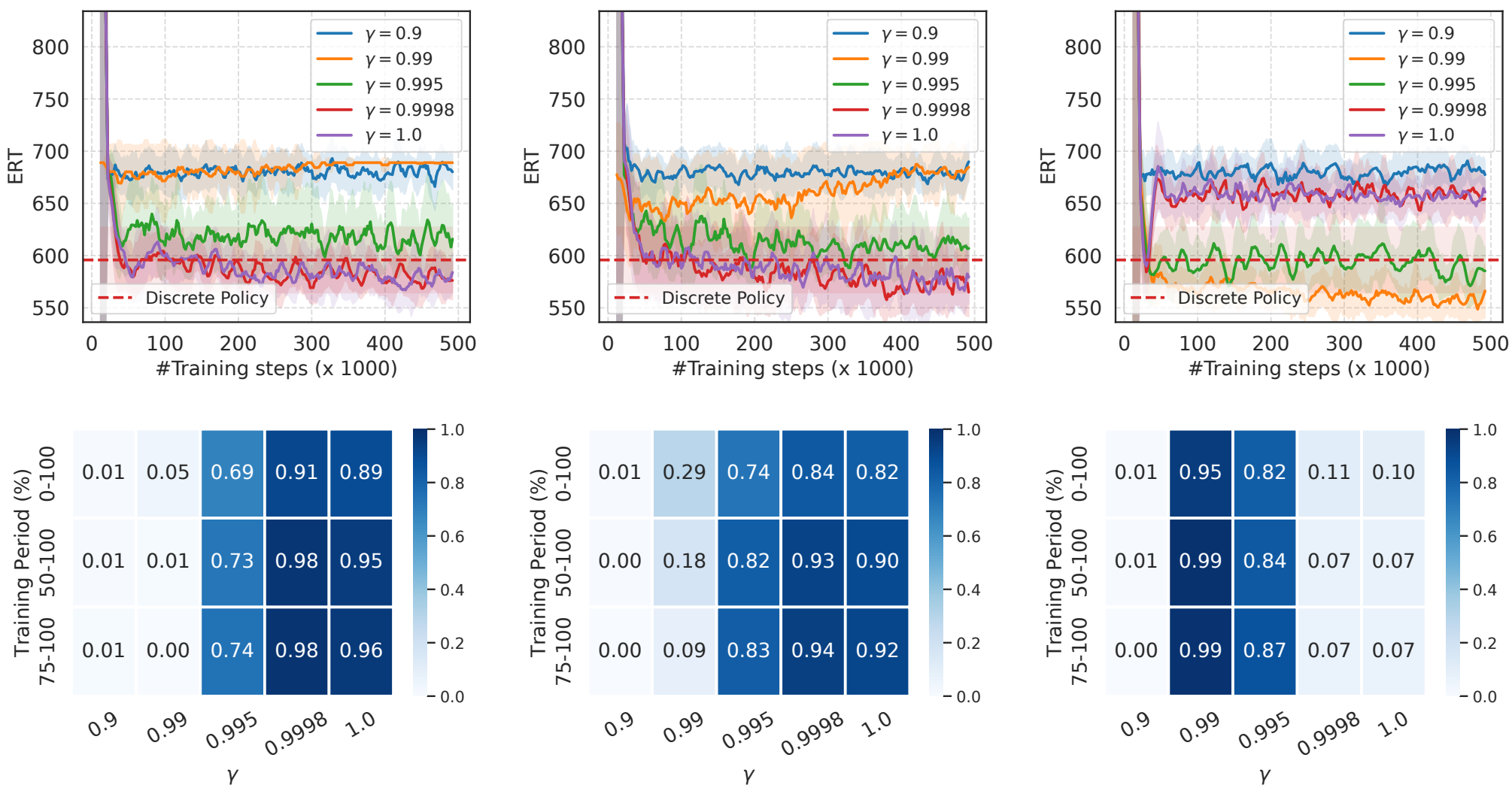

Fig. 14. Learning curves (top) and HR (bottom) of DDQN using naïve (left), scaled (middle), and adaptive shifted (right) reward functions in problem size of 100 across various discount factors.

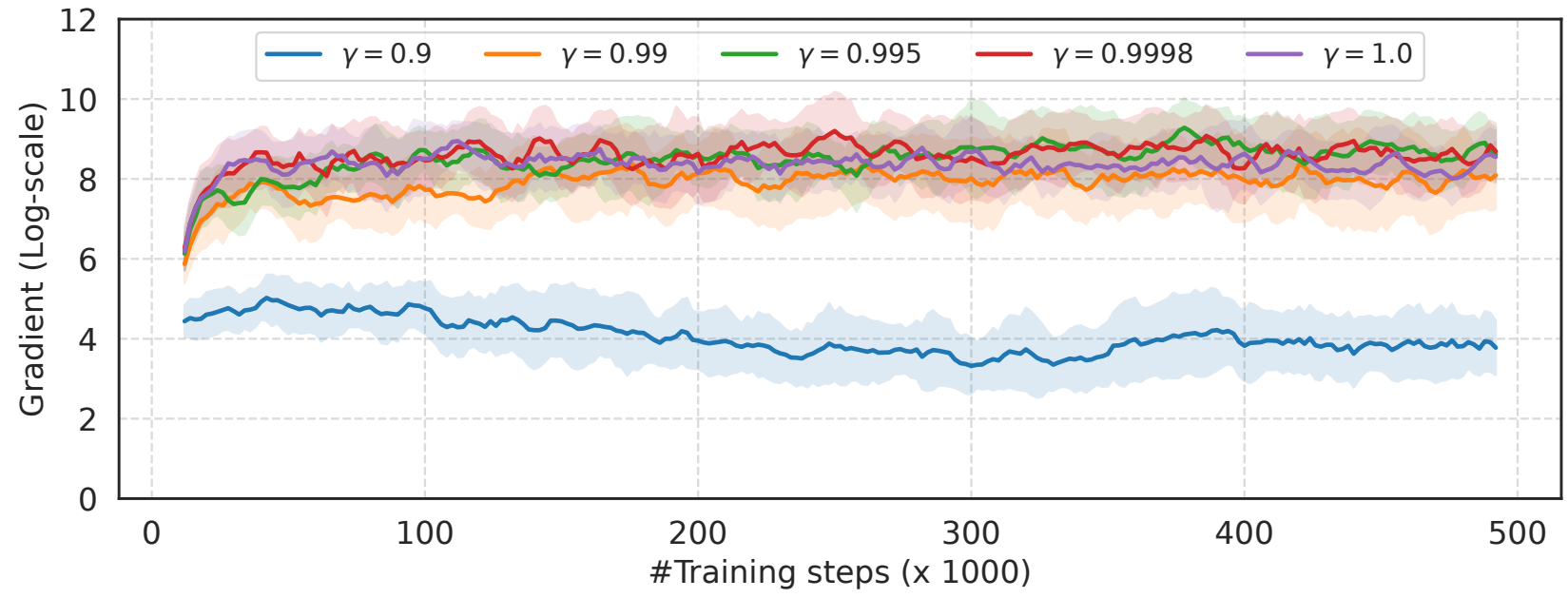

Fig. 15. Tracking the global $L_2$ norm of gradients across all $Q$-online network parameters in DDQN (in natural log-scale), incorporating adaptive reward shifting and different discount factor values, within a problem size of 100. The solid lines indicate the average, while the shaded areas show the variance across 10 RL runs.

Figure 14 (top) reveals contrasting behaviors between naïve/scaled reward functions and reward shifting for $n = 100$. While undiscounting effectively addresses learning divergence issues for naïve and scaled reward functions, it exhibits significantly degraded performance when combined with reward shifting (as indicated by the red and purple solid lines in Figure 14, top-right). The hitting rate analysis (Figure 14, bottom) shows that adaptive shifted reward function with $\gamma = 0.99$ slightly more stable than the naïve reward function with $\gamma = 1$ across all stages of training. However, this advantage of shifting mechanism diminishes as the setup approaches undiscounting ($\gamma = 1$).

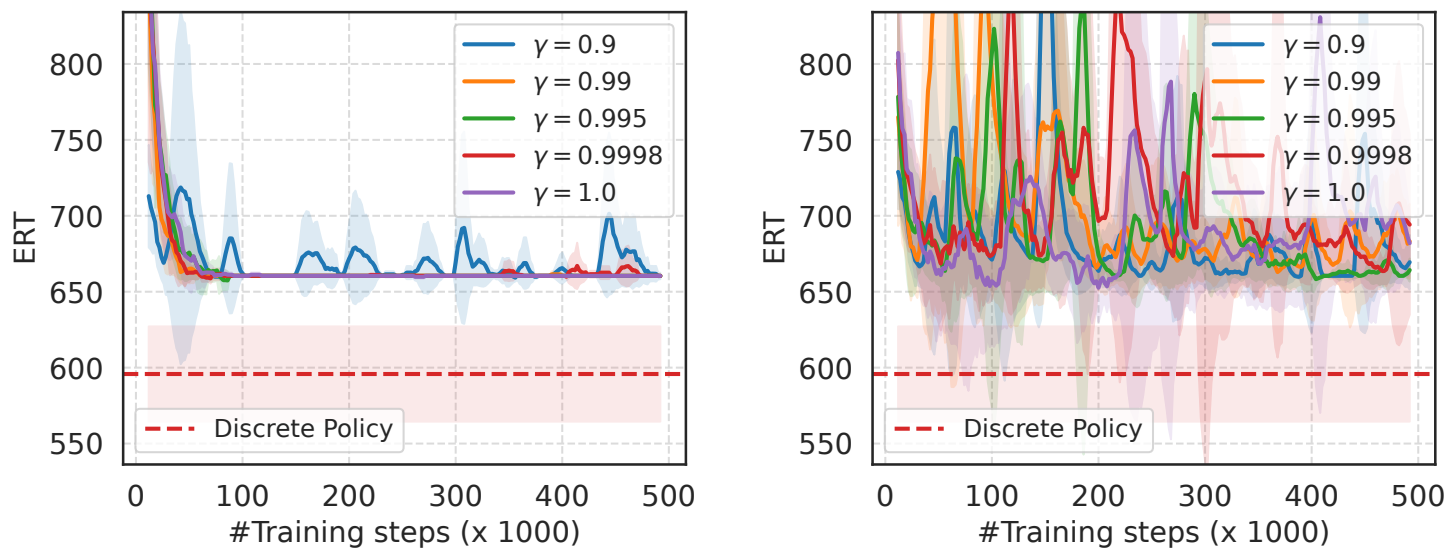

Fig. 16. Learning curves of PPO using naïve (left) and scaled (right) reward functions in problem size of 100 across various discount factors.

We acknowledge that the experiment combining reward shifting with $\gamma = 1$ contradicts the theory-guarantee of reward shifting, as noted in Remark 1 of [Sun et al. 2022], where the shifted $Q$-values are bounded by a gap of $\left|\frac{b}{1-\gamma}\right|$ from using the naïve reward function. This boundary ensures that reward shifting can perform effectively. According to the theory-guarantee, a higher discount factor value leads to a denominator approaching zero, rendering the term of shifting undefined (or approaching infinity). This may cause instability in the $Q$-network learning, potentially leading to an explosion and divergence. We conjecture that this is the reason for the increase in ERT of reward shifting when equipped with a higher discount factor, as shown in Table 3. To verify this assumption, in Figure 15, we capture the global $L_2$ norm of gradients[5] during the training process for a problem size of $n = 100$. The curves can be divided into two groups: standard and high gradients. In the more myopic setting where $\gamma = 0.9$, the gradient operates as usual, which is sensible because the $Q$-values are not significantly shifted from those using a naïve reward function, resulting in a relatively small error in the estimation of $Q$-values. In contrast, all four values of the discount factor $\gamma \in \{0.99, 0.995, 0.9998, 1.0\}$ result in drastically scaled gradients during the first 30,000 training steps. They all then stabilize around $e^8 \approx 3{,}000$, instead of exploding as suggested by the forecast based on the theory-guarantee of reward shifting, particularly in the case of undiscounting (i.e., $\gamma = 1.0$). These observations present a new finding regarding the boundary of reward shifting, indicating that the shifting distance does not always extend to infinity. We conclude that this is beyond the scope of the effectiveness of reward shifting. Additionally, there is only a single setting of the default discount factor ($\gamma = 0.99$) that works well in Table 3, which raises the crucial problem of hyperparameter sensitivity in the reward shifting technique. Although these observations may not fully explain the implicit conflict between reward shifting and the higher value of the discount factor, they offer useful insights for both the evolutionary computation and reinforcement learning communities. One potential direction is for RL practitioners to consider long-horizon planning in the DAC context to explore the limits of reward shaping.

In conclusion, the two challenges of scalability and learning divergence in the OneMax-DAC environment are effectively addressed by two distinct approaches: reward shifting (as discussed in Section 5) and undiscounting (by employing a higher value of the discount factor). Undiscounting provides a simple yet effective solution for DDQN with predictable behavior, though learning remains fairly unstable. Reward shifting emerges as a preferable alternative when learning stability is prioritized.

[5] Calculated as the norm of the loss gradient with respect to all network parameters, mathematically expressed as $\|\nabla_\theta L\|_2 = \sqrt{\sum_i \left(\frac{\partial L}{\partial \theta_i}\right)^2}$, where $\theta_i$ represents the $i$-th trainable parameter and $L$ is the loss function.

Table 4. ERT (± std) normalized by the problem size $n$ across 1000 seeds of the best PPO policies achieved by each combination of {naïve, scaled} reward functions, and discount factors ($\gamma$), across four problem sizes $n \in \{50, 100, 200, 300\}$. Blue: RL outperforms theory-based. Bold: best non-optimal ERT.

| | | **ERT(↓)** | | | |
|---|---|---|---|---|---|
| | | $n = 50$ | $n = 100$ | $n = 200$ | $n = 300$ |
| $\pi_{\texttt{cont}}$ | | 5.448(1.53) | 5.826(1.18) | 6.167(0.97) | 6.278(0.84) |
| $\pi_{\texttt{disc}}$ | | 5.480(1.49) | 5.934(1.28) | 6.244(0.98) | 6.298(0.84) |
| *Reward* | $\gamma$ | | | | |
| Naïve | 0.9 | 5.554(2.04) | 6.611(1.82) | 7.890(2.13) | 8.514(2.03) |
| | 0.99 | 5.534(1.99) | 6.670(1.97) | 7.806(2.09) | 8.531(2.12) |
| | 0.995 | 5.554(2.04) | 6.257(1.58) | 7.608(1.78) | 8.401(1.86) |
| | 0.9998 | 5.492(1.97) | 6.539(1.75) | 7.623(1.95) | 8.399(1.91) |
| | 1.0 | 5.554(2.04) | 6.643(1.92) | 7.647(1.74) | 8.491(2.11) |
| Scaled | 0.9 | 5.554(2.04) | 6.622(1.84) | 7.545(1.68) | 8.123(1.66) |
| | 0.99 | 5.526(1.94) | 6.401(1.69) | 7.406(1.63) | 8.041(1.73) |
| | 0.995 | 5.427(1.91) | 6.073(1.48) | 7.168(1.36) | 7.838(1.61) |
| | 0.9998 | 5.127(1.63) | 6.011(1.39) | 7.110(1.49) | 8.075(1.36) |
| | 1.0 | **5.020(1.50)** | 5.969(1.36) | 6.855(1.28) | 8.186(1.33) |
| $\pi_{\texttt{opt}}$ | | 4.928(1.42) | 5.313(1.10) | 5.609(0.89) | 5.750(0.75) |

**PPO Results.** Unlike DDQN's success with undiscounting and reward shifting, PPO's fundamental challenges persist across all discount factor configurations. Figure 16 shows no improvement with higher discount factors compared to default settings. This evidence indicates that our hypothesis primarily apply to value-based RL methods.

The detailed results of PPO with two designed reward functions are presented in Table 4. The combination of PPO with the original reward function continues to yield suboptimal policies, while reward scaling still results in slow learning convergence. It is not surprising that PPO does not work in all small-to-medium problem sizes, except for a few fortunate times where a scaled reward function is combined with larger discount factors in the smallest problem size. These observations pave the way for a promising direction in investigating other aspects of PPO combined with a scaled reward function.

In order to inspect the shortcomings of PPO in long-horizon episode scenarios, we refer to the analyses by Li et al. [2022]. In their work, they applied PPO to find the optimal solution in the Travelling Salesman Problem (TSP) and also acknowledged that PPO should be applied under the concept of undiscounting. They highlighted the inherent challenge of PPO lies in the high variance of trajectory's return. To verify this challenge in OneMax-DAC, we collect the reward for each step of thousands of trajectories rolled out by the discrete theory-derived policy. We then calculate the variances of these trajectories, which have the same length. As shown in Figure 17 (left), the variances in the default case of a discount factor of 0.99 fluctuate less and remain consistently around 20, while the variances in the case of undiscounting ($\gamma = 1$) exponentially increase. Similar to Li et al. [2022], we admit the technique of generalized advantage estimation (GAE) [Schulman et al. 2015], which is employed to mitigate the bias-variance trade-off in policy-based RL training. More concretely, there is a hyperparameter called $\lambda_{\text{GAE}} \in [0, 1]$ that implements the GAE concept; e.g., a lower value of $\lambda_{\text{GAE}}$ reduces variance but increases bias. Despite our efforts to minimize the value of $\lambda_{\text{GAE}}$, it remains ineffective. The authors in [Li et al. 2022] proposed a custom PPO algorithm that aims to capture the variance alongside the average value by incorporating an additional output head within the value network of PPO. We also experimented with this custom PPO in our OneMax-DAC environment, but it still does not seem to be helping at all.

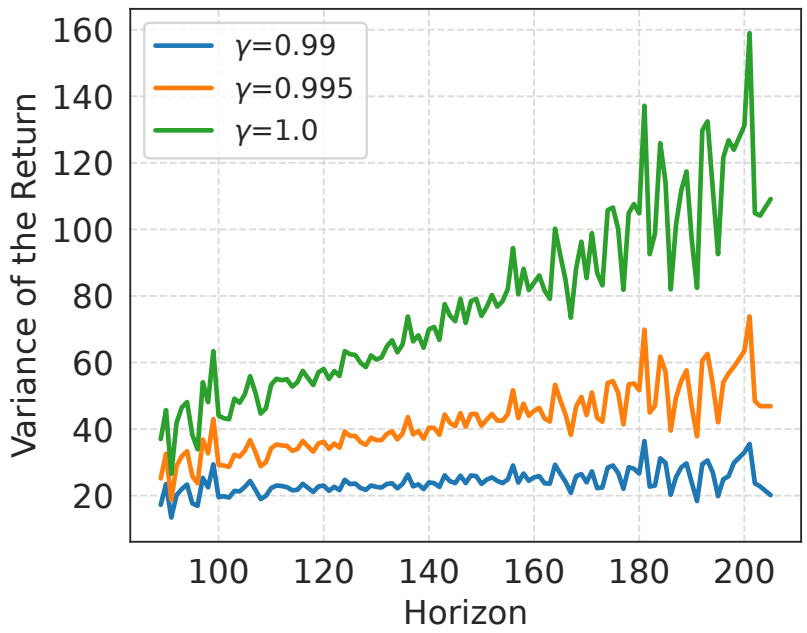

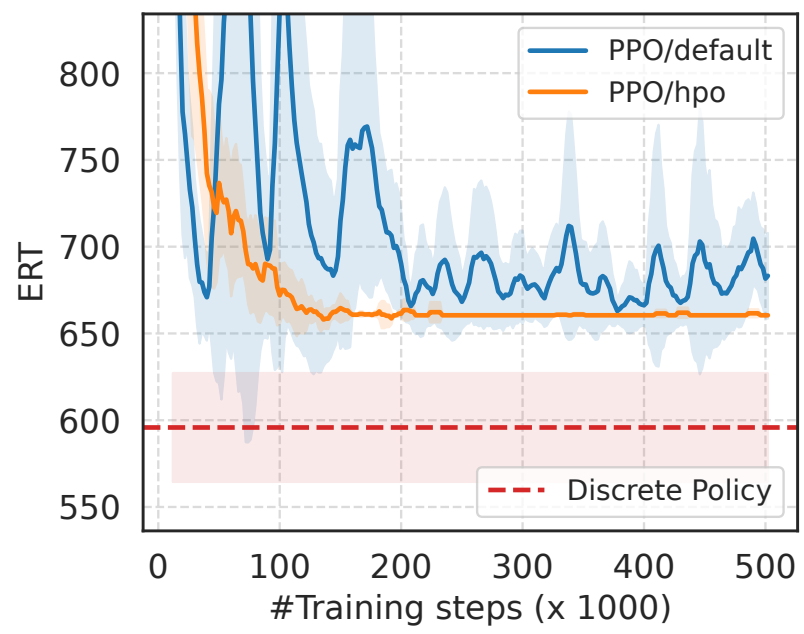

Fig. 17. Left: Variance of returns for three discount factors, $\gamma \in \{0.99, 0.995, 1.0\}$, with a problem size of 100. Right: Learning curves of PPO using a scaled reward function in a problem size of 100, comparing default hyperparameters (blue) and an optimized set of hyperparameters (orange) (refer to Section 7). The shaded regions represent variances across multiple RL runs.

**Discussion.** While undiscounting performs effectively in DDQN, PPO's inherent challenges remain unresolved across all configurations. We attribute these persistent issues to fundamental differences in training strategies between value-based and policy-based RL approaches. Policy-based methods directly estimate gradients of expected returns using sampled trajectories, making them highly susceptible to environmental variance. In contrast, value-based algorithms learn value functions that estimate expected returns for specific state-action pairs, providing indirect policy guidance that is less affected by return estimation variability.

Our analysis reveals that PPO's challenges stem from more than just discount factor selection-the fundamental hyperparameter configuration may be suboptimal for the OneMax-DAC environment. While reward scaling shows promise in PPO (as evidenced by the blue highlights in Table 4), the learning instability suggests that aggressive updates from poorly tuned hyperparameters may be undermining performance. This observation motivates a systematic investigation of hyperparameter optimization for PPO, focusing on the most critical parameters that could address the learning instability and slow convergence issues identified in our experiments, as presented in the next section.

## 7 Hyperparameter Optimzation for PPO in OneMax-DAC

While reward scaling in PPO demonstrates potential (as shown in Table 4), the learning instability suggests that suboptimal hyperparameter configurations may be causing overly aggressive updates that undermine performance. Additionally, given the persistent difficulties with PPO observed in Sections 4.2 and 6, which could not be resolved through straightforward entropy regularization or discount factor adjustments, we investigate whether systematic hyperparameter optimization can address these fundamental challenges.

The success of hyperparameter optimization (HPO) in reinforcement learning depends critically on identifying which parameters to tune, as optimization budgets are inherently limited [Adkins et al. 2024; Becktepe et al. 2024; Eimer et al. 2023; Patterson et al. 2024; Shala et al. 2024; Wan et al. 2022]. Based on our empirical observations and theoretical understanding of PPO's challenges in the OneMax-DAC environment, we identify three critical hyperparameter categories:

Table 5. Comparison of PPO hyperparameter search spaces, Stable Baselines defaults, and optimal configurations identified by HPO.

| Hyperparameter | Search Space | Default | HPO |
|---|---|---|---|
| Learning rate | log(interval(0.00001, 0.01)) | 0.0003 | 0.00001 |
| Mini-batch size | {16, 32, 64, 128} | 64 | 128 |
| Training epochs | range[5, 20] | 10 | 10 |
| Entropy coefficient | interval(0, 0.5) | 0.0 | 0.115 |
| GAE | interval(0.6, 0.99) | 0.95 | 0.705 |
| Clip range | interval(0, 0.5) | 0.2 | 0.442 |
| Discount factor | interval(0.8, 1.0) | 0.99 | 0.986 |

Table 6. ERT and AUC (± std) normalized by problem size $n$, over 1000 seeds of the best PPO policies using the scaled reward function and two distinct hyperparameter configurations. Bold: best non-optimal ERT.

| | $n = 50$ | | $n = 100$ | | $n = 200$ | | $n = 300$ | |
|---|---|---|---|---|---|---|---|---|
| | ERT(↓) | AUC(↓) | ERT(↓) | AUC(↓) | ERT(↓) | AUC(↓) | ERT(↓) | AUC(↓) |
| $\pi_{\texttt{cont}}$ | 5.448(1.53) | - | 5.826(1.18) | - | 6.167(0.97) | - | 6.278(0.84) | - |
| $\pi_{\texttt{disc}}$ | 5.480(1.49) | - | 5.934(1.28) | - | 6.244(0.98) | - | 6.298(0.84) | - |
| *Hyperparameter* | | | | | | | | |
| Default | 5.526(1.94) | 0.181(0.09) | 6.401(1.69) | 1.160(0.09) | 7.406(1.63) | 2.703(0.59) | 8.041(1.73) | 2.754(0.29) |
| HPO | **5.358(1.84)** | 0.176(0.01) | 6.623(1.92) | 0.802(0.02) | 7.510(1.75) | 2.170(0.03) | 8.269(1.82) | 2.752(0.04) |
| $\pi_{\texttt{opt}}$ | 4.928(1.42) | - | 5.313(1.10) | - | 5.609(0.89) | - | 5.750(0.75) | - |

- Learning Dynamics: the slow and unstable learning patterns observed in previous sections suggest that the learning rate and clip range may be set too aggressively. In addition, PPO's training with relatively few epochs and small mini-batch sizes may contribute to insufficient policy updates.
- Exploration-Exploitation Balance: the tendency for PPO to converge prematurely to suboptimal policies (Figure 8) indicates inadequate exploration. Entropy regularization through the entropy coefficient $\beta$ becomes crucial for maintaining appropriate exploration levels throughout training.
- Bias-Variance Tradeoff: as established in Section 6, the OneMax-DAC environment necessitates a higher discount factor to address the planning horizon problem, leading to high variance in policy gradient estimation, as shown in Figure 17 (left). Consequently, careful tuning of both the GAE parameter $\lambda_{\text{GAE}}$ and the discount factor $\gamma$ is essential to manage variance while preserving learning effectiveness.

**Search Spaces.** Based on our analysis earlier, we define the following search spaces for critical hyperparameters as shown in Table 5. These ranges are informed by our previous observations and established best practices in RL hyperparameter optimization [Eimer et al. 2023], while being sufficiently broad to capture potentially beneficial configurations outside standard defaults.

**Evaluation Protocol.** We track the learning curve and adopt the combined sum of normalized ERT and AUC[6] as the tuning objective. For each proposed hyperparameters configuration, we conduct 3 independent RL training runs on the medium-sized problems ($n \in \{100, 200, 300\}$), with the final score computed as the average performance across multiple runs.

[6]Given the use of the multi-fidelity mechanism, we normalize the AUC by the evaluation count to ensure fair comparisons across varying fidelities.

**HPO Framework.** We employ Hypersweeper [Eimer and Benjamins 2024], a hydra interface [Yadan 2019] for ask–and–tell[7] hyperparameter optimization. Among available black-box optimizers including random search (RS), differential evolution (DE) [Storn and Price 1997], Bayesian optimization (BO) [Jones et al. 1998], SMAC [Hutter et al. 2011] and multi-fidelity extension like Hyperband [Li et al. 2018] (e.g., DEHB [Awad et al. 2021], BOHB [Falkner et al. 2018] and SMAC-HB [Lindauer et al. 2022b]), we prefer SMAC-HB based on its demonstrated advantage in [Eggensperger et al. 2021]. We implement Successive Halving (SH) with a minimum training budget set to 40,000 steps, and a maximum budget of 1 million steps, with a reduction factor $\eta = 2$. This setting ensures that in each bracket stage, the number of configurations is halved while the budget per configuration is doubled, allowing for efficient resource allocation toward promising hyperparameter combinations. Altogether, we evaluate 43 distinct configurations among a budget of 500 trials. For policy evaluations, we employ 10 CPU cores in parallel to expedite the process. Overall, the computationally intensive HPO tuning requires approximately one week to complete.

**PPO-HPO Analyses.** The optimal hyperparameter values for PPO are presented in Table 5. We assessed this configuration across four problem sizes, with the results detailed in Table 6. Overall, HPO does not appear to significantly enhance PPO training performance. This is not surprising for the easiest problem of $n = 50$, where PPO supported by HPO can discover policies that yield ERTs superior to those of the theory-derived policy. For larger problem sizes, however, there is no improvement in PPO policy quality compared to the default hyperparameters; in fact, performance tends to decline. Nonetheless, the AUCs indicate that the training processes become more stable. This suggests a complex interaction between hyperparameters: some enhance exploration but compromise learning stability, while others aim to stabilize learning. As shown in Figure 17 (right), comparing default and tuned hyperparameters, HPO seems to contribute to stabilizing PPO learning rather than to discovering optimal policies.

## 8 Scalability Analysis

Among the two deep-RL algorithms discussed previously, DDQN demonstrates effective performance across all small to medium problem sizes. Consequently, we extend our experiments using DDQN to encompass larger problem sizes $n \in \{500, 1000, 2000\}$ to evaluate its scalability. In Section 5, we have seen that the magnitude of the bias should increase as the problem size increases. Instead of having to tune the bias for each new problem size, we adopt the adaptive shifting bias idea proposed in Section 5.2. Additionally, since the problem sizes are significantly larger, we increase the RL training budget to 1.5 million steps, while keeping the same DDQN settings as in previous experiments. Our objective is to thoroughly analyze the optimal combinations of reward functions and discount factors, as illustrated in Table 3 including: {naïve reward, $\gamma = 1.0$}, {scaled reward, $\gamma = 1.0$}, and {adaptive shifted reward, $\gamma = 0.99$}. We refrain from considering the combination of reward shifting and scaling, as we have observed no advantage over using reward shifting alone. Furthermore, we compare the RL policies with the best policies obtained using the IRACE-based cascading tuning method proposed in [Chen et al. 2023].

In Table 7, we present two metrics including: ERT and number of training time steps required to first surpass the discretized theory-derived baseline $\pi_{\mathtt{disc}}$. While our proposed adaptive reward shifting mechanism shows marginal improvements over other approaches with high discount factors in small- to medium-sized problems, its performance becomes significantly more dominant as the problem size increases. More specifically, the shifting mechanism achieves highly competitive ERT and consistently outperforms the discretized theory-derived baseline $\pi_{\mathtt{disc}}$ across all three problem sizes. It also surpasses IRACE on $n = 500$ but performs worse than IRACE on the larger problem sizes ($n = 1000$ and $n = 2000$), even when statistical tests are used. However, it is important to note that IRACE's tuning budget is substantially larger than our RL training budget. For instance, on $n = 500$, each iteration of IRACE's cascading process consumes 5,000 episodes, amounting to at least $5{,}000 \times 400 = 2{,}000{,}000$ time steps. There are 9 iterations in total, which results in an 18-million time step tuning budget. The tuning budget increases to 75 millions and 308 millions for $n = 1000$ and $n = 2000$, respectively (compared to the 1.5 million budget of RL). The ERT reported in Table 7 reflects the best result IRACE found at the end of this extensive tuning process.

[7]The ask–and–tell interface in hyperparameter optimization separates the generation of candidate hyperparameters (ask) from reporting their evaluation results back to the optimizer (tell).

Table 7. ERT (± std) normalized by the problem size $n$ across 1000 seeds of the best DDQN policies achieved by each combination of {naïve, scaled, adaptive shifted} reward functions, and discount factors ($\gamma$), across four problem sizes $n \in \{500, 1000, 2000\}$. #Steps indicates when the RL agent first achieves a policy outperforming the theory-derived policy across 10 RL runs. Bold: best non-optimal ERT.

| | | $n = 500$ | | $n = 1000$ | | $n = 2000$ | |
|---|---|---|---|---|---|---|---|
| | | ERT(↓) | #Steps($\times 10^6$) | ERT(↓) | #Steps($\times 10^6$) | ERT(↓) | #Steps($\times 10^6$) |
| $\pi_{\texttt{cont}}$ | | 6.474(0.67) | - | 6.587(0.53) | - | 6.681(0.39) | - |
| $\pi_{\texttt{disc}}$ | | 6.543(0.71) | - | 6.701(0.55) | - | 6.821(0.42) | - |
| *Reward* | $\gamma$ | | | | | | |
| Naïve | 1.0 | 6.243(0.71) | 0.108 | 6.720(0.58) | 0.476 | 7.193(0.47) | 0.176 |
| Scaled | 1.0 | 6.661(0.77) | 0.034 | 6.734(0.61) | 0.024 | 6.676(0.46) | 0.144 |
| Adaptive Shifted | 0.99 | **6.216(0.70)** | 0.030 | 6.491(0.56) | 0.032 | 6.664(0.43) | 0.034 |
| IRACE | | 6.261(0.68) | 2002 | **6.422(0.52)** | 20046 | **6.376(0.38)** | 159439 |
| $\pi_{\texttt{opt}}$ | | 5.876(0.65) | - | 6.017(0.48) | - | 6.093(0.36) | - |

To better assess the sample efficiency of each approach, we measure the number of training time steps required to first surpass the discretized theory-derived baseline $\pi_{\texttt{disc}}$. This metric, shown in the #Steps column of Table 7, reveals that DDQN with reward shifting requires several orders of magnitude fewer time steps than IRACE across all three problem sizes, highlighting DDQN's strong advantage in sample efficiency.

## 9 Conclusion

Applying reinforcement learning to dynamic algorithm configuration (DAC) is challenging and typically requires extensive domain expertise. Throughout our systematic investigation of examining DDQN and PPO on the OneMax benchmark, we encountered profound learning bottlenecks. For the evolutionary computation community seeking to investigate reinforcement learning agents in dynamic configuration contexts, our journey offers a methodological reflection on overcoming these uncovered hurdles.

A core lesson emerged in our primary problem formulation. We discovered that framing the DAC problem with a discrete action space provides a robust, simplified starting point that eases the exploration burden. Within this framework, value-based methods like DDQN demonstrated superior sample efficiency and stability. Conversely, policy-based methods like PPO required extreme caution; they consistently suffered from learning instability and premature stagnation, which is a fundamental issue that even the extensive hyperparameter optimization could not resolve.

Reward design is a crucial element in shaping the environment for DAC. We observed that naïve reward signals, which directly map to optimization objectives like minimizing runtime, frequently fail and induce severe numerical instability as problem dimensions scale. While normalizing rewards somewhat mitigated this numerical issue, it did not cure learning stagnation, where agents became safely but suboptimally trapped (e.g., always selecting the smallest population size). We encountered that this was essentially an issue of under-exploration. By implementing an adaptive reward shifting strategy that applies a calculated negative bias based on reward distribution statistics, we successfully incentivized the agent to explore high-performing policies without requiring manual, instance-specific tuning.

Furthermore, we uncovered a critical, often-overlooked mismatch regarding the planning horizon. Standard discount factors (e.g., 0.99) strictly limit the agent's effective planning horizon, which inherently misaligns with the long, extended optimization runs typical in DAC environments. By adopting undiscounted learning (a discount factor of 1.0), working together with naïve reward function, we successfully resolved this horizon coverage mismatch for value-based reinforcement learning. Our experiments also highlight that automated hyperparameter optimization cannot solve underlying issues; HPO can stabilize a poorly tuned setup but cannot correct fundamental algorithmic mismatches, such as PPO's inherent variance issues.

Although our study shows that DDQN works well with the DAC framework of $(1+(\lambda,\lambda))$-GA solving the OneMax problem, a critical open question is how this deep reinforcement learning method and debugging approach will perform on non-linear, multi-modal, or deceptive fitness landscapes. A direction for future work is to investigate the generalization of the RL mechanisms investigated in our study, such as adaptive reward shifting and undiscounted learning ($\gamma = 1.0$) to address the challenges of under-exploration or long planning horizons in DAC environments. Another important direction to explore is on expanding the current theory-inspired benchmarks with richer state representations, such as population or solution profiling, problem identification, or landscape analysis, to enable the RL agent to perceive the landscape's ruggedness before adjusting the algorithm's parameters. Another potential future direction for reward design is to leverage the robustness of large language models (LLMs) for the task of automatic reward design [Huang et al. 2026; Ma et al. 2023; Sun et al. 2025]. Finally, beyond the model-free RL approaches examined in this study, we plan to explore model-based RL, which remains largely under-investigated in DAC and may offer promising avenues for improving the learning's sample efficiency.

## Acknowledgments

The project is financially supported by the European Union (ERC, "dynaBBO", grant no. 101125586), by ANR project ANR-23-CE23-0035 Opt4DAC, and by an International Emerging Action funded by CNRS Sciences informatiques. The work used the Cirrus UK National Tier-2 HPC Service at EPCC funded by the University of Edinburgh and EPSRC (EP/P020267/1) and the supercomputer at MeSU Platform (https://sacado.sorbonne-universite.fr/plateformemesu). Tai Nguyen acknowledges funding from the St Andrews Global Doctoral Scholarship programme. André Biedenkapp acknowledges funding through the research network "Responsive and Scalable Learning for Robots Assisting Humans" (ReScaLe) of the University of Freiburg. The ReScaLe project is funded by the Carl Zeiss Foundation. This publication is based upon work from COST Action CA22137 "Randomised Optimisation Algorithms Research Network" (ROAR-NET), supported by COST (European Cooperation in Science and Technology).

## Supplementary Material for

## "Deep Reinforcement Learning for Dynamic Algorithm Configuration: A Case Study on Optimizing OneMax with the (1+($\lambda$,$\lambda$))-GA"

In this supplementary material, we provide a proof of the shifting gap in optimistic initialization. Next, we detail the coefficient tuning process for the proposed adaptive reward shifting and compare it with the mechanism introduced in [3]. Finally, we demonstrate the policy behaviors across different approaches for various problem sizes.

# A Reward Shifting

The theory of reward shaping was introduced in [2], where the authors presented the concept of learning another MDP model, denoted as $\mathcal{M}'$, instead of the original $\mathcal{M}$. The new $\mathcal{M}'$ is defined as $(\mathcal{S}, \mathcal{A}, \mathcal{T}, \mathcal{R}')$, where $\mathcal{R}' = \mathcal{R} + F$ is the reward set in $\mathcal{M}'$. The trained agent in $\mathcal{M}'$ would also receive a reward of $R(s, a, s') + F(s, a, s')$ when executing the action $a$ to transition from state $s$ to $s'$. They defined a potential-based shaping function following Theorem 1 in [2] $\mathcal{F} : \mathcal{S} \times \mathcal{A} \times \mathcal{S} \mapsto \mathbb{R}$ and a real-value function $\Phi : \mathcal{S} \mapsto \mathbb{R}$:

$$F(s, a, s') = \gamma\Phi(s') - \Phi(s), \tag{1}$$

A transformation from the optimal action-value function $Q_{\mathcal{M}}$ in $\mathcal{M}$ to $\mathcal{M}'$ satisfies the Bellman equation:

$$Q^*_{\mathcal{M}'}(s, a) = Q^*_{\mathcal{M}}(s, a) - \Phi(s) \tag{2}$$

and the optimal policy for $\mathcal{M}'$:

$$\pi^*_{\mathcal{M}'}(s) = \arg\max_{a \in \mathcal{A}} Q^*_{\mathcal{M}'}(s, a) = \arg\max_{a \in \mathcal{A}} Q^*_{\mathcal{M}}(s, a) - \Phi(s) \tag{3}$$

Sun et al. [4] defined the potential-based function $F$ in Equation (1) as a constant bias $F(s, a, s') = b$, thus $\mathcal{R}' = \mathcal{R} + b$ with $b \in \mathbb{R}$, the formula in Equation (1) simplifies to:

$$F(s, a, s') = \gamma\Phi(s') - \Phi(s) = (\gamma - 1)\phi \tag{4}$$

in the case where $F$ is guaranteed to be a constant, thus the potential function $\Phi(s)$ must also be constant $\phi$, then:

$$\Phi(s) = \Phi(s') = \phi = \frac{b}{\gamma - 1} \tag{5}$$

and the Equation (3) becomes (see also the Remark 1 in [4]):

$$\begin{aligned} \pi^*_{\mathcal{M}'}(s) &= \arg\max_{a \in \mathcal{A}} Q^*_{\mathcal{M}'}(s, a) \\ &= \arg\max_{a \in \mathcal{A}} Q^*_{\mathcal{M}}(s, a) - \frac{b}{\gamma - 1} \\ &= \arg\max_{a \in \mathcal{A}} Q^*_{\mathcal{M}}(s, a) + \frac{b}{1 - \gamma} \end{aligned} \tag{6}$$

as the additional bias $b$ that does not depend on the chosen action leads to maximizing the action-value function $Q_{\mathcal{M}'}$ which is equivalent to maximizing the original $Q_{\mathcal{M}}$. The constant $\left|\frac{b}{1-\gamma}\right|$ represents the difference between the altered and the original state.

# B Adaptive Reward Shifting

## B.1 Coefficient Tuning

We initially hypothesize that the bias should be proportional to the average reward: $b^-_{\text{a}} = \alpha\,\bar{R}$, where $\alpha$ is a scaling coefficient. However, this simple formulation fails to generalize effectively across problem sizes because reward ranges scale exponentially with problem dimensions while maintaining different distribution shapes. To achieve better generalization while avoiding explicit dependence on problem size $n$ (which may not be readily available in other DAC applications), we incorporate reward distribution quartiles that naturally capture the exponential scaling behavior. Our final adaptive bias formula incorporates both the reward mean and the ratio of quartiles:

$$b^-_{\text{a}} = \alpha\,\bar{R}\,\frac{Q_1}{Q_3}\,\frac{1}{Q_{\text{base}}}, \tag{7}$$

the term $Q_{\text{base}} = \frac{Q_1(n=100)}{Q_3(n=100)}$ serves as a normalization constant based on our reference problem size of 100. The quartile ratio $\frac{Q_1}{Q_3}$ captures the exponential scaling behavior across problem sizes, while $Q_{\text{base}}$ ensures the bias magnitude remains within the effective range observed for our reference case. The scaling coefficient $\alpha$ controls the overall magnitude of the bias adjustment. This coefficient is determined empirically by comparing the resulting bias values with the optimal range of fixed bias for the problem size of 100, the optimal value of $\alpha$ should be $\frac{1}{12}$. Substituting the empirically determined values, this simplifies to:

$$b^-_{\text{a}} = 0.0052\,\bar{R}\,\frac{Q_1}{Q_3} \tag{8}$$

**Table 1: ERT (± std) normalized by the problem size $n$ of optimal, theory-derived and RL-based DAC policies with two adaptive shifting mechanisms combined with two designed reward functions across seven problem sizes. Blue: RL outperforms theory-based. Bold: best non-optimal ERT. Underline: not statistically significantly different from the best ERT (according to Bonferroni corrected paired t-tests with a confidence level of 99%).**

| | ERT(↓) | | | | | | |
|---|---|---|---|---|---|---|---|
| | $n=50$ | $n=100$ | $n=200$ | $n=300$ | $n=500$ | $n=1000$ | $n=2000$ |
| $\pi_{\texttt{cont}}$ | 5.448(1.53) | 5.826(1.18) | 6.167(0.97) | 6.278(0.84) | 6.474(0.67) | 6.587(0.53) | 6.681(0.39) |
| $\pi_{\texttt{disc}}$ | 5.480(1.49) | 5.934(1.28) | 6.244(0.98) | 6.298(0.84) | 6.543(0.71) | 6.701(0.55) | 6.821(0.42) |
| Shifting($Q_2$) | 5.087(1.41) | 5.511(1.18) | 5.794(0.93) | 6.080(0.83) | **6.080(0.66)** | **6.459(0.56)** | 6.800(0.45) |
| Scaling_Shifting($Q_2$) | 4.992(1.42) | **5.383(1.22)** | 5.943(1.05) | 6.220(0.98) | 7.186(0.94) | 7.047(0.66) | 7.928(0.62) |
| Shifting($Q_1/Q_3$) | **4.919(1.44)** | 5.444(1.19) | **5.723(0.93)** | **5.990(0.85)** | 6.216(0.70) | 6.491(0.56) | **6.664(0.43)** |
| Scaling_Shifting($Q_1/Q_3$) | 4.927(1.41) | 5.409(1.19) | 5.723(0.93) | 6.103(0.93) | 6.769(0.88) | 6.589(0.57) | 6.889(0.44) |
| $\pi_{\texttt{opt}}$ | 4.928(1.42) | 5.313(1.10) | 5.609(0.89) | 5.750(0.75) | 5.876(0.65) | 6.017(0.48) | 6.093(0.36) |

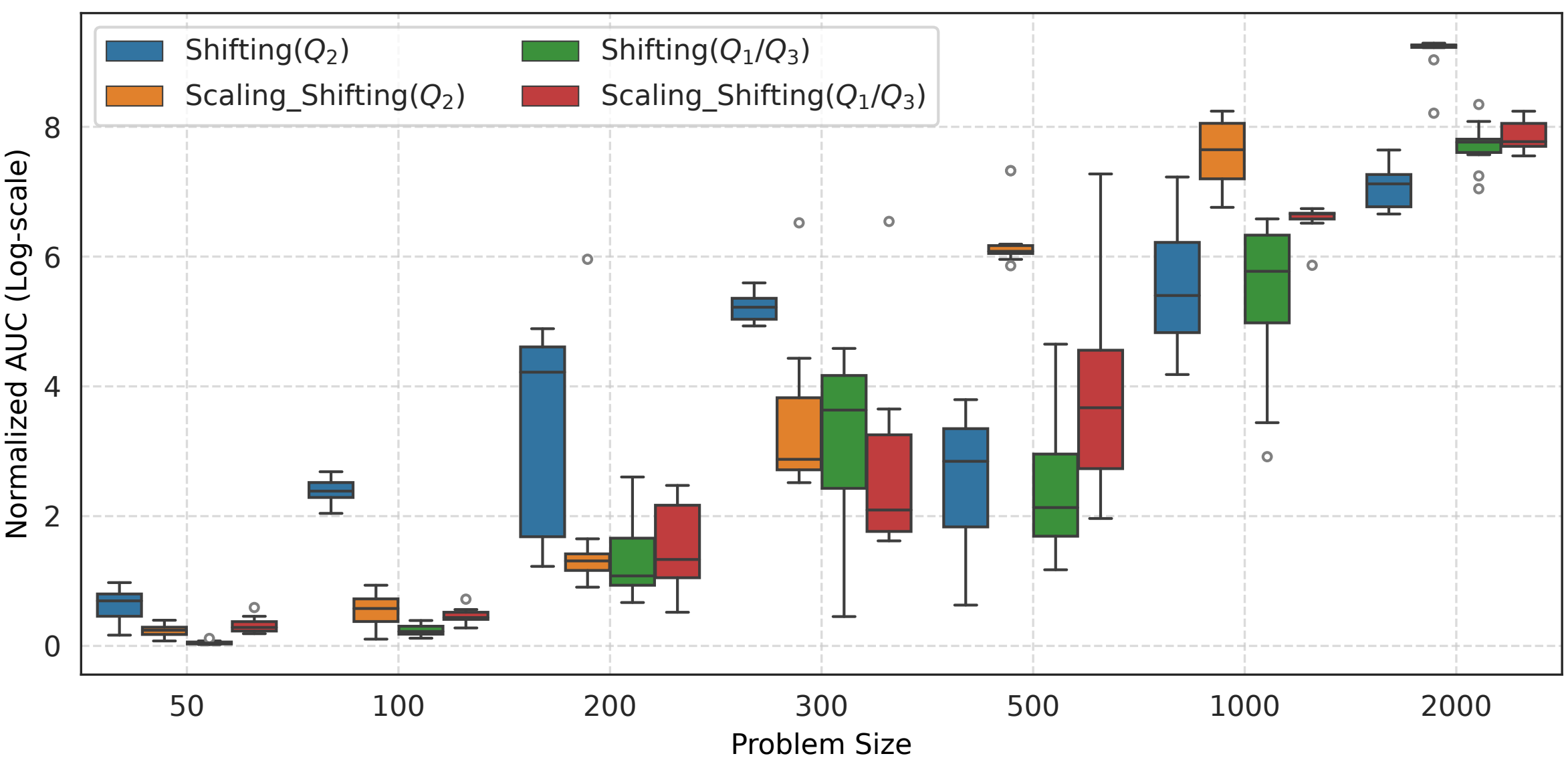

**Figure 1: AUC results normalized by the problem size $n$ for two adaptive shifting mechanisms combined with two designed reward functions across seven problem sizes.**

## B.2 Discussion

We've observed that our proposed new formula for adaptive reward shifting is more general and performs competitively better than the median-dependent formula in [3]. In this section, we compare the two adaptive shifting mechanisms in terms of ERT and AUC. As shown in Table 1, when applying a naïve reward function, our proposed adaptive shifting method based on the ratio of the first and third quartiles ($Q_1/Q_3$) that competitively outperforms the previous approach that relies on a median-dependent ($Q_2$) mechanism, especially in the largest problem size tested. Furthermore, experiments with larger problem sizes reinforce our earlier observations, indicating that the combination of scaling and shifting mechanisms does not yield promising results. In terms of learning capabilities, the green boxes in Figure 1 generally outperform the blue boxes, representing that our proposed adaptive shifting enhances learning stability more effectively than the prior method [3].

# C Policy Comparison

We present in Figure 2 the policies across 4 approaches for problem sizes $n \in \{100, 200, 300\}$ and an additional comparison with IRACE for problem sizes $n \in \{500, 1000, 2000\}$. For RL, we present the best policies selected from the top-5 policies in the evaluation phase during training. These policies incorporate the reward shifting mechanism into the original reward function. Similar to [1], we plot $\lambda$ values only for $f(x) \geq n/2$, as this is the most relevant region, given that a random initial solution typically has a fitness around $n/2$.

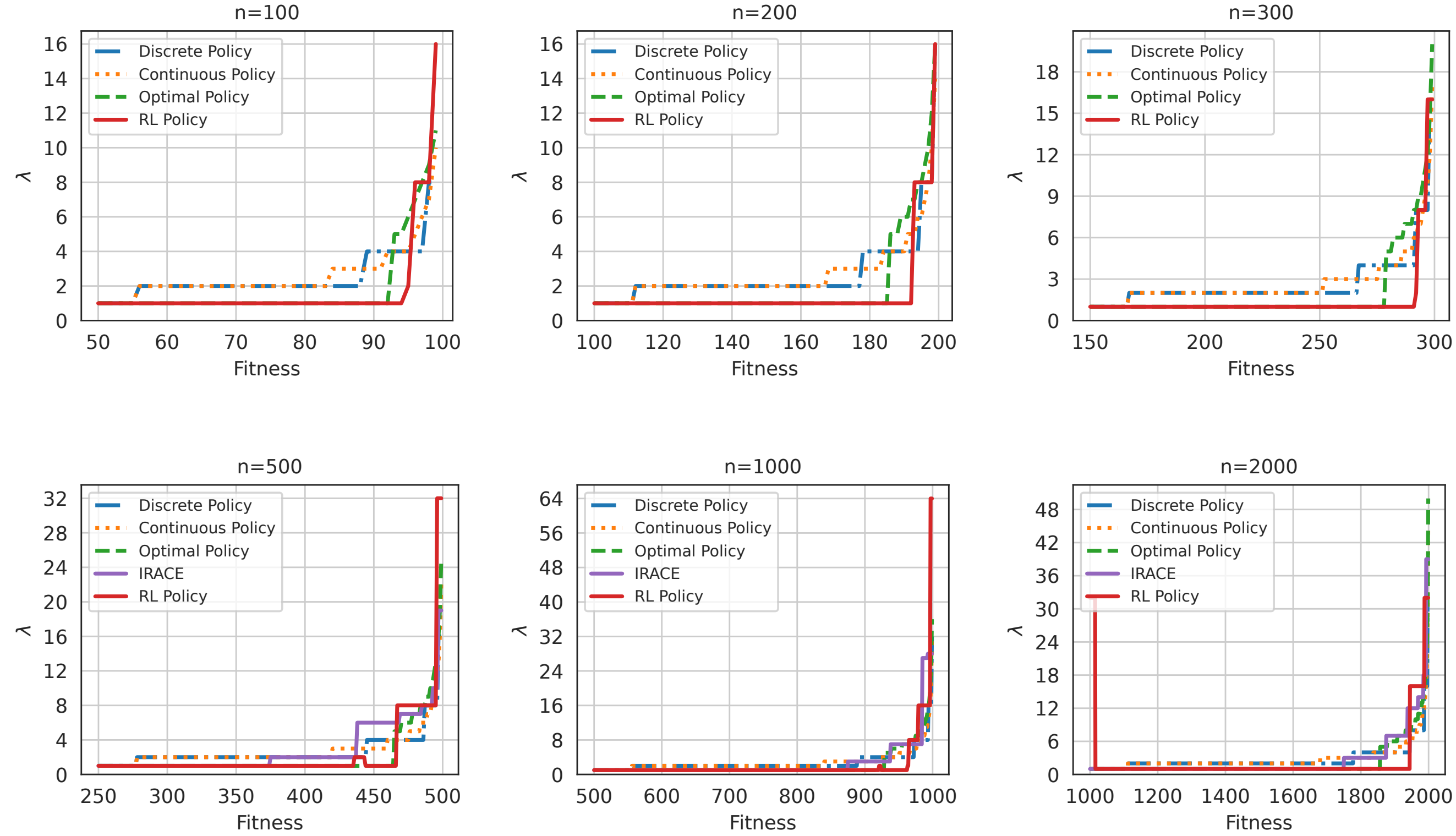

**Figure 2: Policies of the theory-derived $\pi_{\texttt{cont}}$, its discretized version $\pi_{\texttt{disc}}$, the optimal policy $\pi_{\texttt{opt}}$, IRACE-based policies, and RL, which denotes our best-trained DDQN for $f(x) \geq n/2$.**